# Knowledge Graphs: The Future of Data Integration and Insightful Discovery

**Saher Mohamed, Kirollos Farah, Abdelrahman Lotfy, Kareem Rizk, Abdelrahman Saeed, Shahenda Mohamed, Ghada Khouriba, Tamer Arafa**

School of Information Technology and Computer Science (ITCS), Nile University Giza, Egypt

{Smfarag; k.saleh; Alotfy; K.ayman, A.Saeed, shatem; GhadaKhoriba; Tarafa}@nu.edu.eg

**Abstract**. Knowledge graphs are now used as an eficient way of repre-senting and connecting information on various concepts. It can be applied to reasoning, question answering or knowledge base completion tasks. Knowledge graphs are an organized way of presenting data, with one feature being that data points are linked to each other. There are also several advantages to using knowledge graphs for academic researchers and scientists working with big and diverse data in various fields. The ability to put together a large amount of information from various fields into a single database enables researchers to come up with new questions that do not have to belong to a single discipline. For instance, an academic in medicine who focuses on analyzing the social consequences of an illness might use data from journals, patients' records, census data and economic statistics to identify new information on vulnerable com-munities. Making these connections explicit can be useful in coming up with new ideas for research or in identifying potential interdisciplinary projects. It is imperative that academia incorporate the use of knowl-edge graphs in order to propel studies in times of big and complex data. A knowledge graph creates a web of data points which are nodes and edges that join the nodes together. Nodes represent single entities, while edges depict the connections between those things. This structure of the network enables knowledge graphs to create relationships between the different data points and unveil them to be easily navigated, compre-hended and utilized for diverse purposes. This graph structure also en-ables representation of relations and connections between different nodes or vertices such as, persons, companies, ideas or occasions. Traditional databases and knowledge representations struggle to effectively capture the wide range of relationships inherent in unstructured data sources like text. Knowledge graphs address this challenge by providing an intercon-nected semantic framework that can represent diverse types of entities and their associated attributes and interrelationships. There are several strategies that organizations can adopt when developing their knowl-edge graphs, and this depends on the types of data sources to be used, the kind of entities to be incorporated in the graph and the number of users and uses of the graph. One approach is to begin with such care-fully selected seed data as knowledge bases, and then use techniques such as named entity recognition, relationship extraction, and entity disam-biguation on free text sources to identify new nodes and edge for the graph. Knowledge graphs can also be used to aggregate different data sources of an organization into a common knowledge model. They also vary in the extent to which they are built manually or automatically, as well. Defining the knowledge domain and evaluating the needs guaran-tees that organizations create graphs designed for the specific purpose and goals. Knowledge graphs offer information gathered from various sources and interconnecting related concepts, making data available for query, reasoning, and inference. The graph structure allows for the ex-ploration of connected entities and the relationships between them that one may not have seen when just looking at the points of data. It in-dicates that they offer context and semantics that enrich applications. This is especially important for tasks that involve context and semantic relations between words, such as entity recognition or relation extrac-tion. Knowledge graphs also enhance the ability of chatbots and virtual assistants to provide more accurate and relevant responses. They can build better answers from the rich semantic information encoded into the data. It is also important to note that information in the knowl-edge graph is not only textual; it can also include data from multimedia files, including images, videos, and audio files. By including rich me-dia in their knowledge graphs, organizations can obtain more extensive information on users' preferences, attitude analysis, and content recom-mendations. Knowledge graphs allow for easy navigation and finding information within the data. Through the representation of nodes and edges and the possibility of users to interact with them, it is possible to reveal new patterns or trends. The use of graphs as a visual repre-sentation provides a way to explore the relationships between related concepts that may not be apparent when looking at just the numbers. Knowledge graphs visually link the nodes, meaning that within the data, the stories are told in a way that makes understanding easier. This pro-motes deeper engagement and communication, as stakeholders utilize these visual narratives to make better decisions. The interactivity in the construction of knowledge graphs and the emergent properties of the explorations allow individuals to find knowledge and wisdom where it would not be apparent, thus helping them make better decisions. How-ever, creating accurate knowledge graphs is not easy even though they are useful. Merging real-world data with different and disordered structures, while maintaining reliability and comprehensiveness, is not straightfor-ward. The new methods such as knowledge graph embeddings and graph neural networks are useful to automate some parts of graph construction and extension. Still, human intervention and step-by-step improvement are likely to be indispensable for reaching higher levels of quality that are necessary for mission-critical applications that rely on the knowledge embedded in these graphs. Knowledge graphs drive functionality in nu-merous applications. In this case, Google used the graph to make search results more accurate and informative by providing additional informa-tion about an entity like a person and their achievements, and other concepts that may be relevant. This gives the searcher an overall picture of what they are looking for without being redirected. Netflix, for exam-ple, uses knowledge graphs to make recommendations to their users. One good example of this is Netflix where it uses the patterns and preferences within a graph in order to recommend

Movies or tv shows based on the search of the user. Knowledge graphs also help with semantic search not only in textual data. The process of knowledge graph construction and maintenance is a promising field that has been recently attracting a lot of attention from researchers and developers, and it is the case when the employment of the information extraction, entity linking, and relation learning methods allows for the creation of knowledge graphs starting from the unstructured data sources. However, the question of how to ob-tain accurate, complete, and consistent knowledge graphs remains one of the most significant hurdles, and the creation of most high-quality knowl-edge graph resources still involves a combination of automated methods, supervision, and curation. Recent advancements have shown a lot of promises in creating automated methods for knowledge graph construc-tion from text through the use of NLP and ML. Tools like SpaCy offer enhanced entity recognition and relationship extraction capacities that can parse vast volumes of text and extract entities, concepts, and their relationships at speed. Language models have also played a role in this area as they are able to solve tasks such as entity linking, taxonomy construction, and knowledge base population. Nevertheless, fully auto-mated approaches are still not well suited for dealing with the ambiguity, complexity, and inevitable mistakes in analyzing unstructured data. The usual practice is to receive final approval from domain specialists and validate the extracted knowledge graphs manually for increased accuracy. Therefore, the majority of the knowledge graph projects have adopted the use of both human and artificial intelligence.
*abstract* environment.

# 1 Introduction

The goal of artificial intelligence (AI) is to mimic human thinking, which involves more than just learning from specialized books. It’s similar to how a doctor combines textbook knowledge with real-life practice. While AI models have made progress in learning from experience, they often miss out on the valuable insights found in structured literature, affecting their explainability and performance. Knowledge graphs (KGs) have become an important tool in this area. They organize human knowledge in a way that’s similar to how books are structured, making it easy for AI to access and use specific information. KGs provide a strong framework to represent various entities and their relationships, making them useful for many fields like social interactions, biological processes, academic references, and transport systems. These graphs are essential for AI systems to understand and operate effectively Hogan et al. (2021).

There are two main types of knowledge graphs: static and dynamic Liang et al. (2022). Static KGs offer a stable foundation of information for AI, much like the base of a building. In contrast, dynamic KGs can adapt and grow with new data and changing contexts, similar to an ever-changing landscape. Building and maintaining these knowledge structures involves collecting and organizing data in a way that AI can use, and continuously updating them with new information. This requires strategic planning to ensure the graphs stay relevant and accurate. This book chapter examines the detailed processes behind creating and man-aging both static and dynamic KGs. It looks at how these graphs are developed, how they evolve, and how they are kept up-to-date. The goal is to find the best ways to manage these extensive information networks so that AI systems remain
effective in a constantly changing environment.

## 2 Background and Terminology

The history of knowledge graphs dates back to the development of semantic networks in artificial intelligence during the early 1960s. These networks repre-sented knowledge as nodes (concepts) and edges (relationships), creating a graph that illustrates the interrelation of concepts. Ontologies emerged in the 1990s, formalizing the representation of a set of concepts within a domain and the re-lationships between them to facilitate knowledge sharing and domain reasoning. At the turn of the millennium, Tim Berners-Lee Berners-Lee et al. (2001)in-troduced the concept of the Semantic Web, an extension of the current web pro-viding well-defined meaning to information, allowing both computers and people to work in cooperation. Linked Data principles were crucial to this vision, linking
data for easy machine processing.

The term ”knowledge graph” gained popularity when Google launched its Knowledge Graph in 2012. This vast network of interlinked descriptions of en-tities aimed to deliver more relevant search results by understanding entity se-mantics.The essence of a knowledge graph, according to a cited work, can be captured as:

“A knowledge graph is a multi-relational graph composed of entities and re-lations that are regarded as nodes and different types of edges, respectively.”

1 illustrates the evolution of technologies and concepts in the field of knowl-edge graphs from the 1960s to the present. It shows a timeline with key mile-stones, starting with semantic networks in the 1960s, followed by ontologies in the 1970s, the introduction of Linked Data principles around the 1980s, and the development of knowledge graphs, with a specific emphasis on the Semantic Web and dynamic knowledge graphs.

Since then, knowledge graphs utilize a static structure that provides a snap-shot of knowledge at a single point in time. However, this approach has limita-tions in representing knowledge that is constantly evolving. Dynamic knowledge graphs address this by enabling the graph structure and information to change over time as new data is incorporated. This adaptive capability of dynamic knowledge graphs allows for more robust and reasoned modeling of knowledge domains that are rapidly changing and interconnected. As additional related data is acquired, the knowledge graph can automatically restructure and up-date itself to represent the most current understanding. Outdated information can be removed or modified while newly discovered relationships and concepts can be added in an ongoing manner. This adaptive capability means dynamic knowledge graphs offer a living knowledge model rather than a static snapshot of information captured at a single point in time. The ability to continuously update knowledge representations has important applications.

Technologies that rely on structured knowledge models to function, such as question-answering systems, conversational agents, and personalized recommen-dation engines, can leverage dynamic knowledge graphs to ensure they utilize the most up-to-date information available. Complex decision-making systems for applications such as guiding autonomous vehicles can also benefit greatly from dynamically evolving knowledge models that adapt in real time to new road con-ditions or situations. Most significantly, dynamic knowledge graphs address a key limitation of static graphs for representing the massive amounts of knowledge in our digital world which is growing at an exponential rate. By facilitating contin-uous evolution and integration of new online information from diverse sources, dynamic knowledge graphs enable knowledge bases to effectively scale and keep up with the updates of data. This makes them essential for modeling knowl-edge in today's fast -changing digital where information is in a constant state of updating. Incorporating dynamic knowledge graphs (KGs) into computer vi-sion applications significantly enhances capabilities in areas such as autonomous vehicles and production line fault detection. For autonomous vehicles, dynamic KGs facilitate real-time learning and adaptation to new road conditions and scenarios, improving decision-making processes and ensuring safer navigation. Dynamic KGs enhance object recognition, ensuring the vehicle's perception sys-tem accurately identifies and understands its surroundings, even as external conditions change. Additionally, the rich semantic representation provided by dynamic KGs aids in contextual understanding, crucial for interpreting complex scenes and making preemptive safety decisions.

In the realm of production line fault detection, dynamic KGs offer continu-ous learning from production data, identifying patterns and anomalies indica-tive of faults, and ensuring the system's anomaly detection remains accurate and relevant as production processes evolve. Dynamic KGs excel in integrating multimodal data from various sources, providing a comprehensive view of the production process and identifying potential issues before they escalate. Climate prediction can benefit from continuous monitoring of environmental variables like temperature, humidity, and pollution levels. By leveraging dynamic KGs, autonomous vehicles and production line fault detection systems and climate change prediction achieve heightened levels of accuracy, reliability, and adapt-ability, ensuring their effectiveness and safety in a constantly changing world.

Fig 2 provides overview of the process involved in constructing and utilizing a knowledge graph from heterogeneous data sources. It highlights the contrast between human and machine methods in transforming raw data into structured knowledge and wisdom. Firstly, the figure categorizes data into three types: structured data, unstructured data, and semi-structured data. Structured data includes text, existing knowledge graphs (KGs), and tables. Unstructured data encompasses PDFs, images, voice, and video. Semi-structured data includes for-mats such as XML, web pages, and JSON. These various data types are essential inputs into the knowledge graph construction process. The different brains in-volved in the process are then depicted. The human brain is responsible for processing information, transforming it into knowledge, and eventually into wis-dom. On the other hand, the machine brain involves data preprocessing and analysis, transforming raw data into useful insights. In the data preprocessing and analysis phase, the machine brain handles the crucial step of converting unstructured and semi-structured data into a format suitable for knowledge ex-traction. The knowledge graph construction process is divided into three main parts: knowledge extraction, knowledge fusion, and knowledge refinement. In the knowledge extraction phase, the system identifies entities within the data (entity extraction), extracts attributes of the identified entities (attribute extraction), and identifies relationships between entities (relation extraction). The knowledge fusion phase involves aligning entities across different data sources (entity align-ment) and linking entities to ensure consistency and integration (entity linking). Finally, the knowledge refinement phase includes classifying entities into prede-fined categories (entity classification), predicting potential relationships between entities (relation prediction), and identifying and addressing anomalies in the data (anomaly detection). The final step involves knowledge graph reasoning and representation, which enables advanced applications like semantic search, data integration, and decision-making. This step ensures that the knowledge graph can be used effectively to provide insights and support complex queries, ultimately facilitating better understanding and utilization of the underlying data.

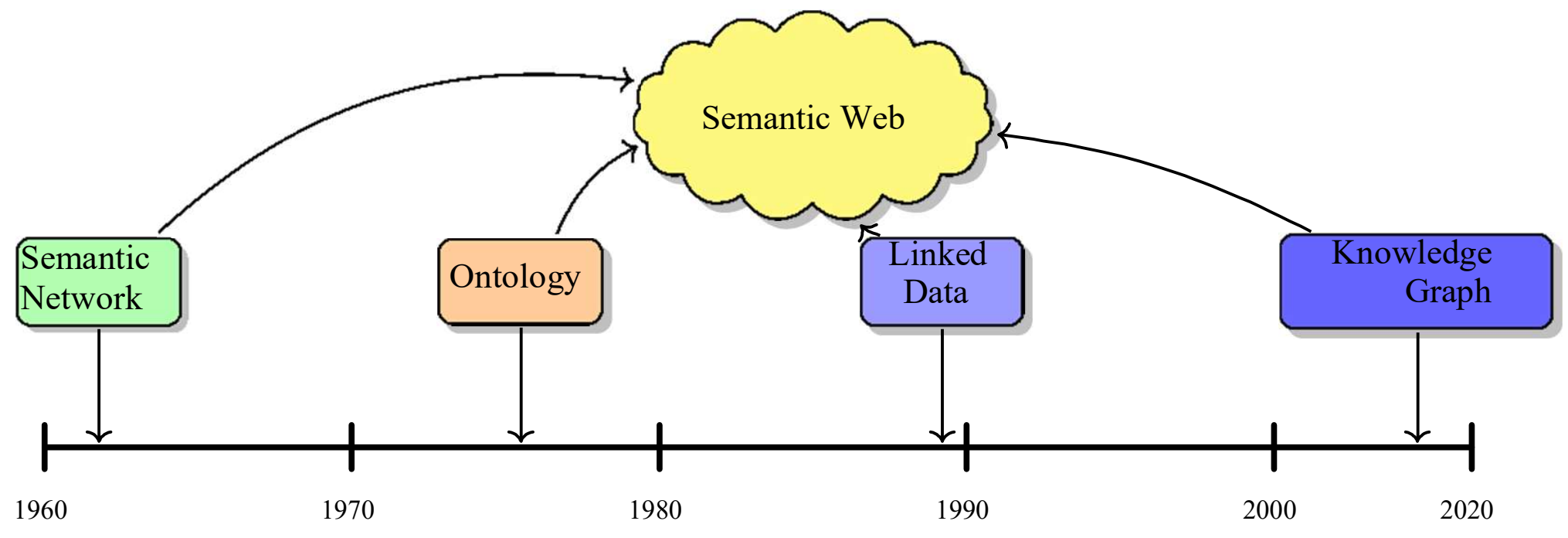

Fig. 1. A timeline illustrating the evolution of technologies and concepts in the field of knowledge graphs, from semantic networks in the 1960s to the Semantic Web and dynamic knowledge graphs.

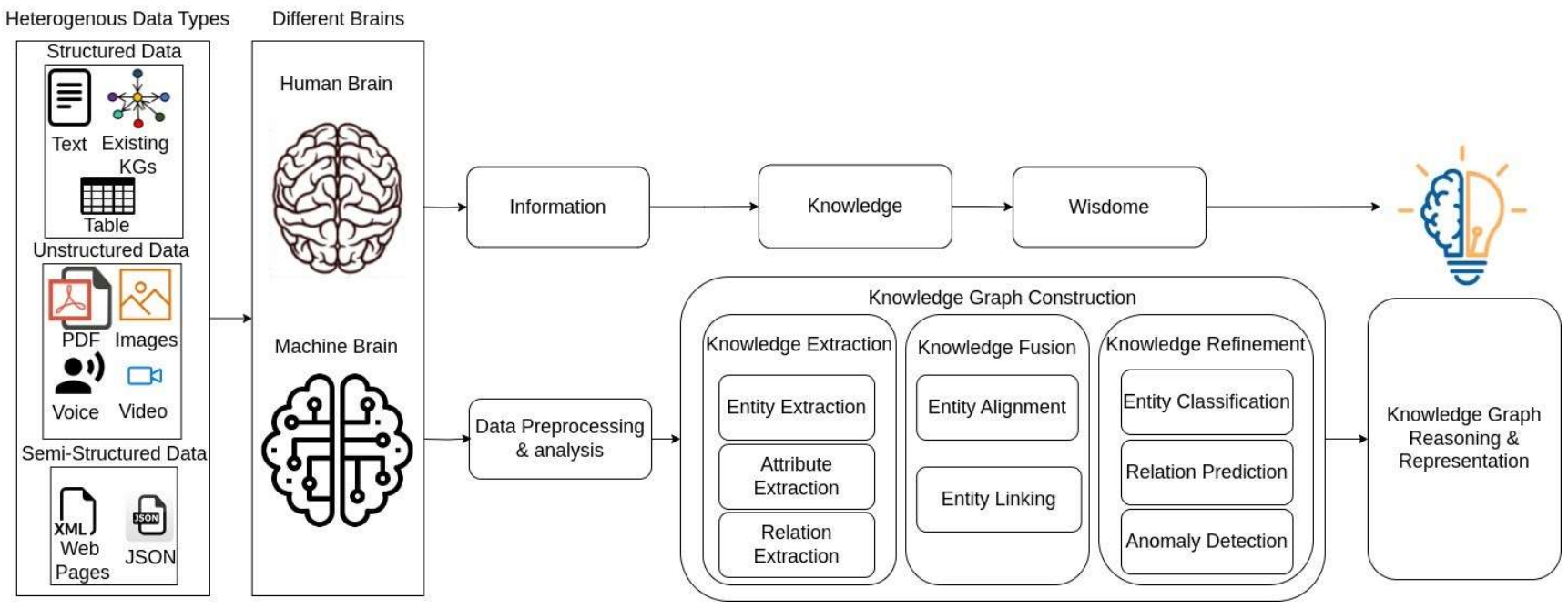

Fig. 2. An overview of the process for constructing and utilizing a knowledge graph from heterogeneous data sources.

## 3 Comparison Between Static Knowledge Graph and Dy-namic Knowledge Graphs

Table 1 provides a detailed comparison between static and dynamic knowledge graphs across various features. Each feature is explained for both static and dynamic knowledge graphs, highlighting their differences in terms of definition, data updates, nature of data, use cases, advantages, challenges, data volume, complexity, examples, versioning, querying, construction methods, construction tools, and main applications.

Table 1: Comparison of Static and Dynamic Knowledge Graphs

| Feature | Static Knowledge Graphs | Dynamic Knowledge Graphs |
|---|---|---|
| **Definition** | A knowledge graph with fixed information that does not change over time. | A knowledge graph that is updated continuously to reflect changes in knowledge over time. |
| **Data Updates** | Rarely or never updated. | Frequently updated with new data. |
| **Nature of Data** | Historical, archival, or unchanging information. | Current, real-time, and rapidly changing information. |
| **Common Use Cases** | Encyclopedic data, lexical databases, and foundational scientific knowledge. | Real-time analytics, social networks, financial markets, and news aggregation. |
| **Advantages** | Stability, reliability, and consistency in data. | Adaptability, relevancy, and timeliness of information. |

Table 1 Continued from previous page

| Feature | Static Knowledge Graphs | Dynamic Knowledge Graphs |
|---|---|---|
| **Challenges** | Can become outdated; may not reflect current knowledge or trends. | Complex to maintain; requires effective data update and versioning mechanisms. |
| **Data Volume** | Typically, less volume compared to dynamic graphs due to the lack of frequent updates. | Potentially large and rapidly growing volume of data due to constant updates. |
| **Complexity** | Generally simpler in structure and maintenance. | More complex due to the need for handling updates and temporal data. |
| **Examples** | DBpedia, YAGO, WordNet. | Twitter graph, Google Knowledge Graph, LinkedIn's Economic Graph. |
| **Versioning** | Not typically necessary. | Often requires version control to track changes over time. |
| **Querying** | Static queries based on a snapshot in time. | Dynamic queries that can handle temporal aspects and changes. |
| **Methods of Construction** | Compilation from reliable sources and manual curation. | Automated collection and integration of data streams, machine learning for pattern recognition and updates. |
| **Construction Tools** | Ontology editors (e.g., Protégé), RDF frameworks (e.g., Apache Jena). | Stream processing engines (e.g., Apache Kafka), dynamic graph databases (e.g., Neo4j with temporal plugins). |
| **Main Applications** | Domain ontologies, educational tools, research databases. | Social media analytics, trend forecasting, adaptive learning systems, and IoT data management. |

## 4 Motivation

There is a significant academic motivation behind the development of dynamic knowledge graphs. The reasoning capability in static knowledge graphs does not change quickly with the addition of new information or the discovery of new linkage. Since knowledge domains online are evolving at such high speed, static graphs do not easily upgrade with updates that understand new relationships or discover them. Dynamic knowledge graphs tackle this challenge by providing an avenue for continual knowledge improvement. Unlike static structures with fixed nodes and edges, dynamic graphs are flexible enough to be modified as new insight appears. The continuous evolution of the graph will help improve inferences made over time. With the concept change and their interconnections due to new incoming data from its source, the graph evolves to increase its power for inference. Dynamic graphs can find new inferences that were not at all possible before by incorporating the latest information coming in from texts, databases, and the web. They can update existing logical rules when knowl-edge evolves. This dynamism in modeling relationships between ideas is a step towards the long-term goals of commonsense reasoning and question-answering systems. Whereas static graphs give only a snapshot of understanding, dynamic graphs can thus harness the potential of deepening insights with the progressive accommodation of new knowledge.

# 5 How can we develop a data integration methodology that effectively combines heterogeneous data sources to con-struct static knowledge graphs, ensuring data reliability and accuracy?(RQ1)

## 5.1 Data Source Diversity and Complexity

Knowledge graphs (KGs) can be categorized into two types: generic knowledge graphs, which cover multiple domains with encyclopedic content such as Wiki-data, YAGO Suchanek et al. (2007), Freebase Bollacker et al. (2008), and DB-pedia Lehmann et al. (2015), and domain-specific knowledge graphs, focused on narrower domains for specific problems or industries Abu-Salih (2021).

KG development typically follows either a top-down or a bottom-up ap-proach. The top-down approach involves defining the ontology or data schema first, followed by knowledge extraction based on this ontology Li et al. (2020a). In contrast, the bottom-up approach starts with knowledge extraction from data, with the ontology of the KG being defined based on the extracted information Li et al. (2020a).

The definition of a knowledge graph's domain is instrumental in the effective identification of data sources and the subsequent determination of data extrac-tion methodologies Yan et al. (2018). This domain delineation can range from broad, encompassing areas such as education Chen et al. (2018), healthcare Li et al. (2020b), and agriculture Chen et al. (2019), to more specific sectors like so-cial media analysis Lian et al. (2017) or autonomous vehicle technologies Luettin et al. (2022).

Upon establishing the domain, the next critical step is the identification of suitable data sources. These sources, varying in structure and format, signifi-cantly influence both the overall development process of the knowledge graph and the selection of appropriate knowledge extraction techniques. Typically, data may be categorized as structured, semi-structured, or unstructured. Structured data, such as those found in relational databases or tabular formats, are char-acterized by a clear and defined structure. Semi-structured data, like XML files, possesses a certain level of organization, though not as rigid as structured data. In contrast, unstructured data, including plain text, lacks a predefined format or structure Weikum et al. (2021).

Data sources for knowledge graphs are diverse, ranging from online ency-clopedias such as Wikipedia Wu et al. (2020), structured databases Chi et al. (2018), semi-structured documents Ryen et al. (2022), to unstructured textual content Yu et al. (2020). Often, a combination of these varied sources is utilized Yan et al. (2020). The choice of data acquisition methods is thus contingent upon both the nature of the data and its source. The selected method must align with the specific data requirements essential for constructing the knowledge graph. As a result, the required data is not only acquired but also appropriately prepared for the ensuing process of knowledge extraction.

In static KGs, the emphasis is on integrating stable, historical data sources. The domain definition is essential in guiding the identification of relevant data sources, which are typically structured or semi-structured or unstructured. Such KGs often utilize a top-down approach, where the ontology directs the data in-tegration process. The data sources for static KGs include structured databases, XML documents, and other well-organized data formats Weikum et al. (2021).

Dynamic KGs, on the other hand, are designed to be adaptable and evolving, integrating data from continually updated sources. These KGs are prevalent in domains where real-time data is vital, like social media analytics or self-driving car technologies. Dynamic KGs often employ a bottom-up approach, with the ontology evolving as new data is integrated. These graphs harness a mix of structured, semi-structured, and unstructured data, reflecting their dynamic nature Weikum et al. (2021).

While both static and dynamic KGs rely on diverse data sources, their key distinction lies in the approach to data integration. Static KGs prioritize histor-ical data with a structured ontology, whereas dynamic KGs focus on real-time data integration and ontological flexibility.

## 5.2 Knowledge Extraction Techniques from Unstructured data

This section focuses only on unstructured data, specifically text.

### 5.2.1 Large Language Model in Knowledge Extraction

Large language models have become invaluable tools for performing knowledge extraction from unstructured text sources. Due to their pre-training on vast amounts of text, LLMs like BERT, GPT-3 and LLMAs have developed strong abilities to analyze language, identify patterns and relationships within text. One of the primary ways LLMs facilitate knowledge extraction is through named entity recognition. By leveraging their linguistic understanding, LLMs can efi-ciently identify entities mentioned in text like people, organizations, locations, products and more. This extracted entity information can be structured into a knowledge base or graph. In addition, LLMs are also capable of performing relationship extraction to understand how different entities might be connected based on analysis of language and contextual cues. This allows modeling rela-tionships between entities as structured triples. Llama 7b Touvron et al. (2023) model has shown great results in consistently extracting triples from unstruc-tured data. The process first involves preparing the unstructured data and for-matting it suitably as input. A prompt is then created to instruct the model to extract subject-predicate-object triples from the unstructured data. By con-structing the prompt in this way to specify triples should be extracted in this sin-gle format, it helps the model to maintain consistent output and avoid potential issues like varying output structures across multiple responses. This optimized approach to structuring both the data preparation step and the triple extrac-tion process with a consistent prompt design allows the large language model to reliably extract structured knowledge in the form of triples from unstructured data in a repeatable manner without changing the output format in each model response as shown in fig 3.

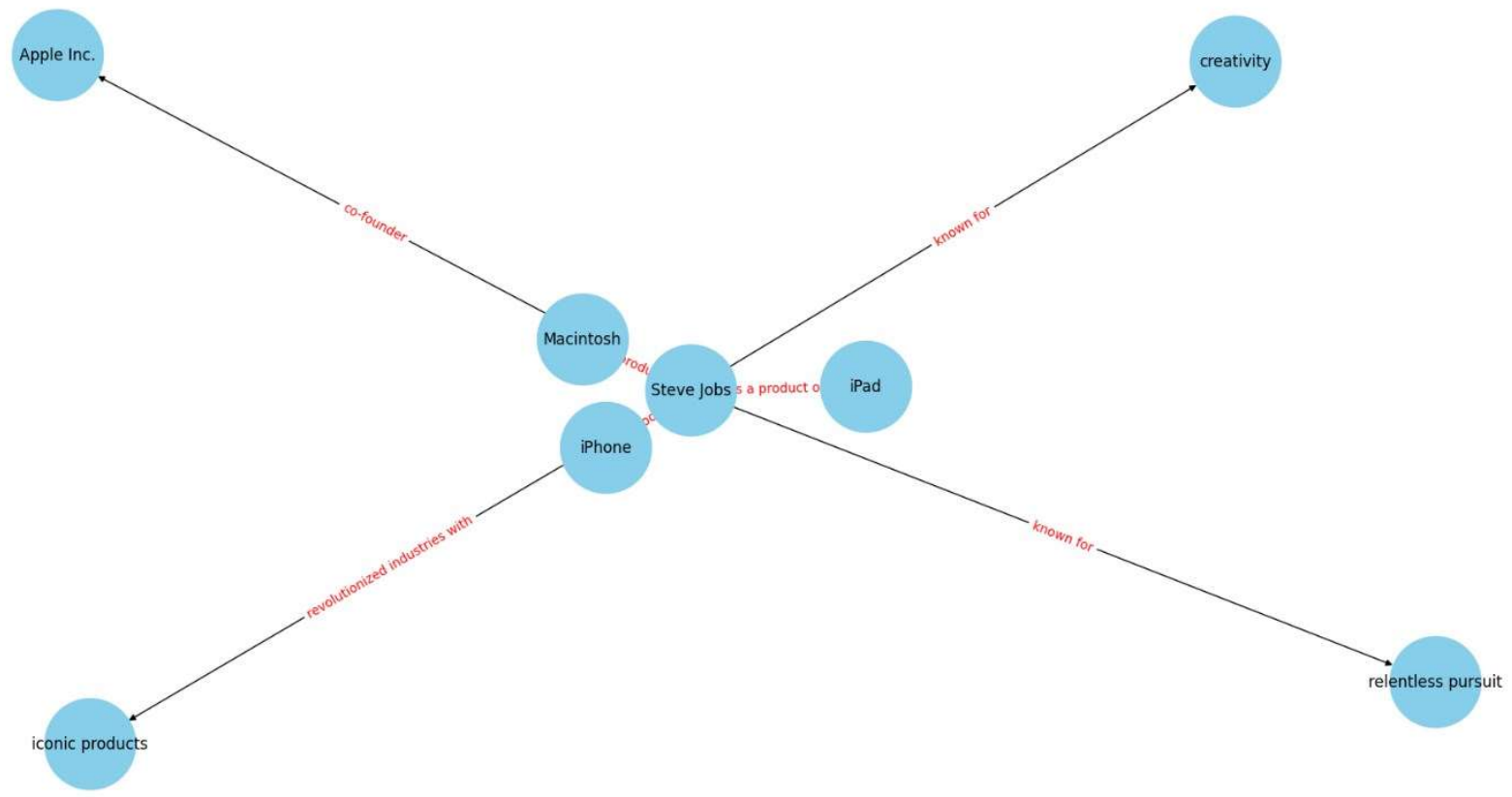

Fig. 3. Knowledge Graph Extracted Using Llama 7b

## 5.3 Natural Language Processing in Knowledge extraction

Sophisticated natural language processing methods are needed to build knowl-edge graphs from unstructured text sources. A significant preliminary process is the pretreatment of input data using linguistic analysis tools. This makes it pos-sible to extract entities and relationships from raw sentences based on syntactic patterns. Many popular NLP libraries, such as spaCy Srinivasa-Desikan (2018), offer strong neural network models that are pre-trained on large corpora and allow the identification of parts of speech, noun phrases, etc., which is necessary for data extraction purposes. After extracting entities and relations, they must be organized into a graph representation with nodes connected to show seman-tic relationships. NetworkX is a great graph library capable of producing the knowledge graph format by turning pre-processed extractions into entities that are nodes and relationships as edges. Textual knowledge can be extracted with expert natural language processing and then modeled programmatically as a net-work structure that powers applications such as question-answering, document retrieval, or semantic search.

As depicted in Figure 4, the initial step involves the creation of a knowledge graph without using the get-entities function. This graph illustrates how predict-ing relations can be directly extracted from text. Subsequently, in Fig 5, a more refined approach is shown where the spaCy model is used for natural language processing and linguistic analysis, loading a pre-trained English language model. The “get − entities” function takes a sentence as input, spaCy tokenizes it, and identifies the entities based on these tokens and their dependencies. The subject entity is the first entity, and the object entity is the second, with valuable entities identified by skipping punctuations and identifying modifiers and compounds. This approach returns a list with pairs of entities, enhancing the accuracy of the knowledge graph.

Furthermore, the “get−relation” function was built to extract the relations from the sentences using the spaCy matcher to identify a sequence of tokens matching a specific pattern. This pattern is specified for extracting the root of a sentence along with optional prepositions, agents, and adjectives. Then, the span of the last match is extracted from the document to return the extracted relation. As shown in Fig 5, this method allows for the creation of a more accurate and detailed knowledge graph.

Finally, Fig 6 demonstrates the knowledge graph with additional predicted relations without using the get-entities function. This figure shows the expanded network, which includes further relationships predicted by the model, providing a comprehensive view of the connections between entities.

The “nx.from−pandas−edgelist” function in the NetworkX library created a directed graph, where the nodes are the entities and the edges are the extracted relations, facilitating the construction of detailed and informative knowledge graphs from unstructured text.

From the Karate Club library, the Deepwalk implementation was initialized with a specific dimension; the nodes in the knowledge graph were relabeled using enumeration and stored in a dictionary. The Deepwalk fitted onto this relabeled graph, generating random walks on the graph and learning node embeddings based on these walks. These embeddings were visualized in a 3D scatter plot. The worde2vec model can also get the embeddings by loading the “word2vec-google-news-300” and requiring us to fetch this model from Gensim API or by using the Node2vec model by fitting the model to apply the Node2vec algorithm on the knowledge graph and also generating some random walks on the KG to training the models and retrieve the embeddings in embedding vector for a specific node.

The spaCy model has a problem in the creativity of defining the entities and the relations because it takes each word in the sentence and defines its entity or review, providing a huge knowledge graph with unusable entities. Then, the logic of extracting entities or relations should be implemented based on the data domain.

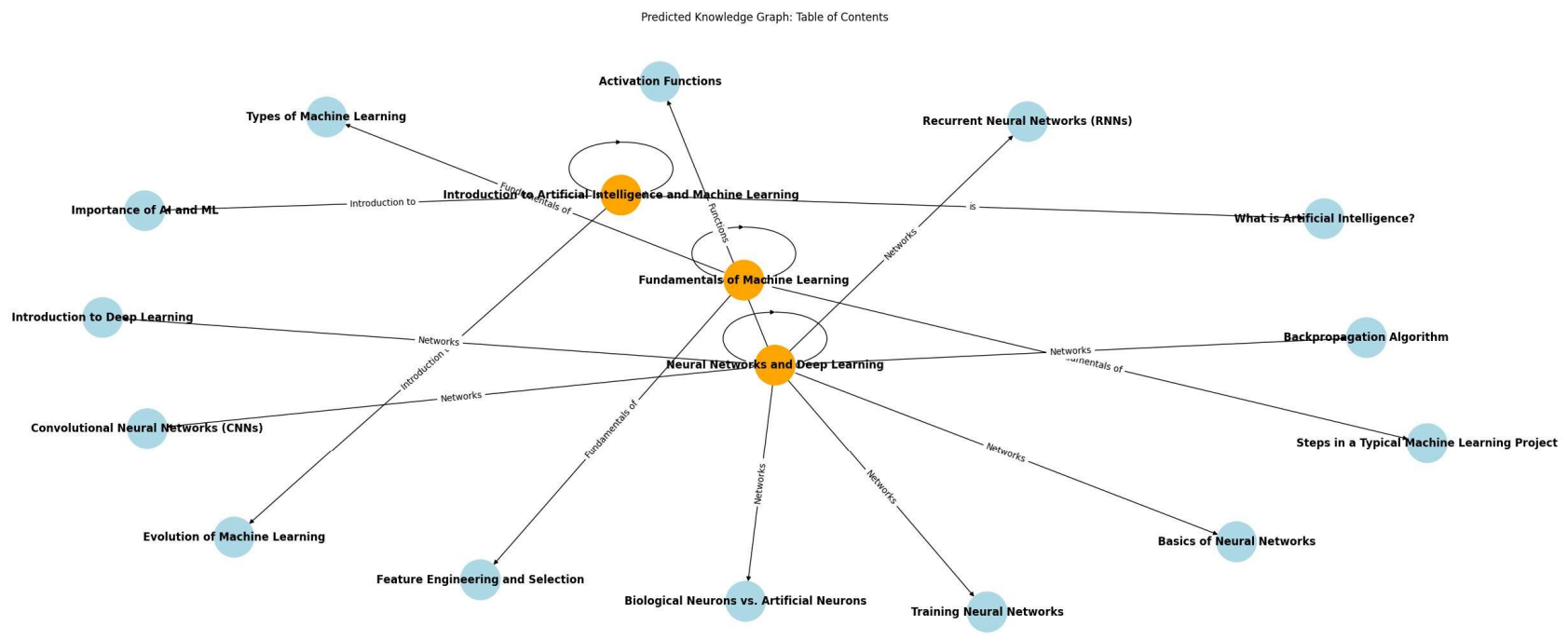

Fig. 4. The knowledge graph with predicting relations without using get-entities.

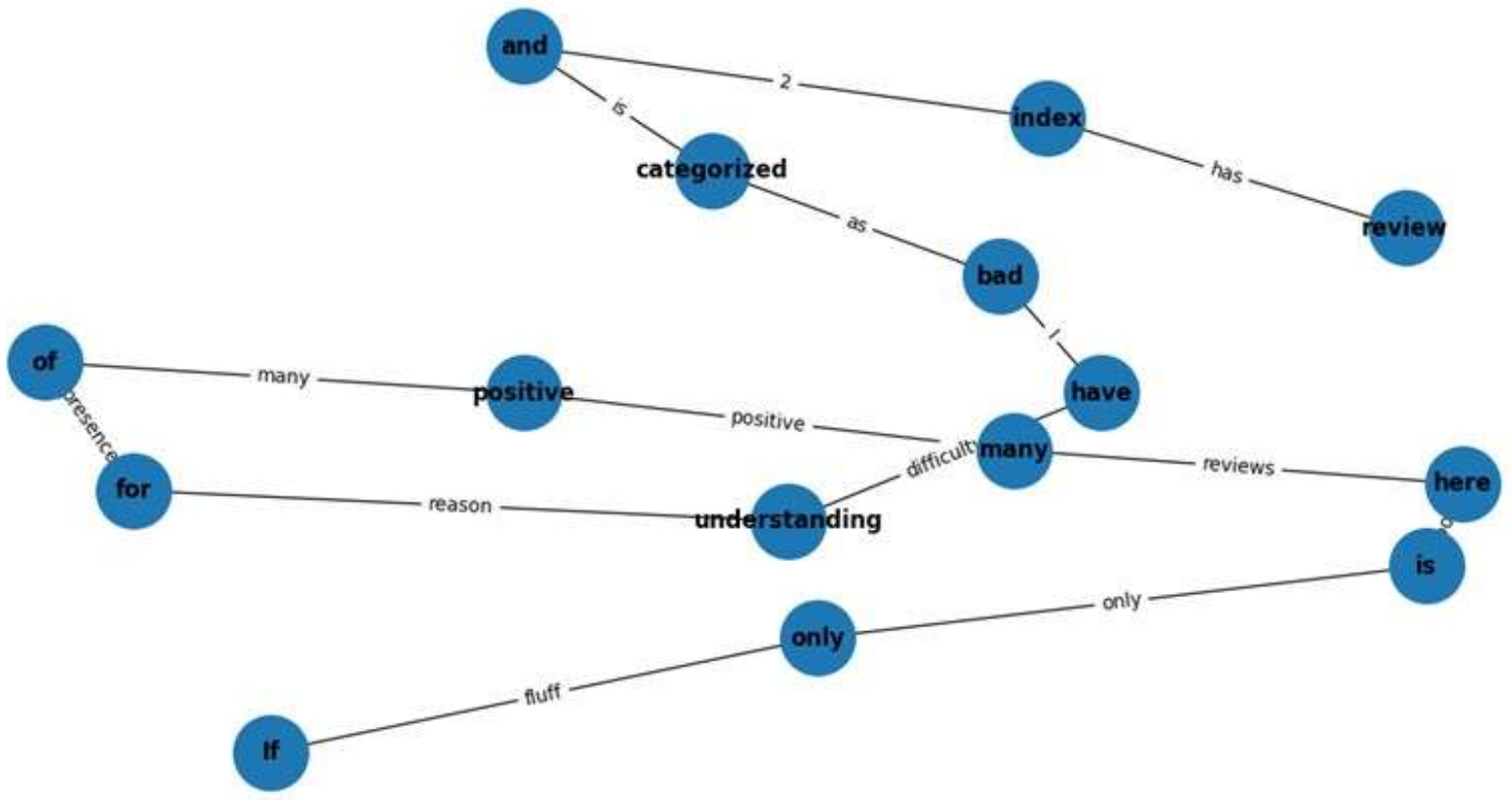

Fig. 5. Sample part of review KG using get-entities and get-relation

# 6 How can we transform a static knowledge graph into a dynamic knowledge graph? (RQ2)

## 6.1 Clustering Techniques in Link Prediction and Adding New En-tities

### 6.1.1 Traditional Clustering

Clustering the entities in the knowledge graph allows groups of similar nodes to be identified, helping to understand the underlying themes or topics repre-sented. A number of algorithms were tested to cluster the entities derived from the Wikipedia sentences. K-means clustering employs centroid-based assignment using node embeddings generated by models like node2vec and DeepWalk. The optimal number of clusters is determined through metrics such as silhouette score. Hierarchical agglomerative clustering builds clusters incrementally using distance-based linkage. Gap statistics estimate the best grouping by comparing within-cluster variance against an expected random distribution. Evaluation of clustering accuracy is done through similarity measures like silhouette scores to group entities with high within-cluster similarity while maintaining the distinc-tion between clusters.

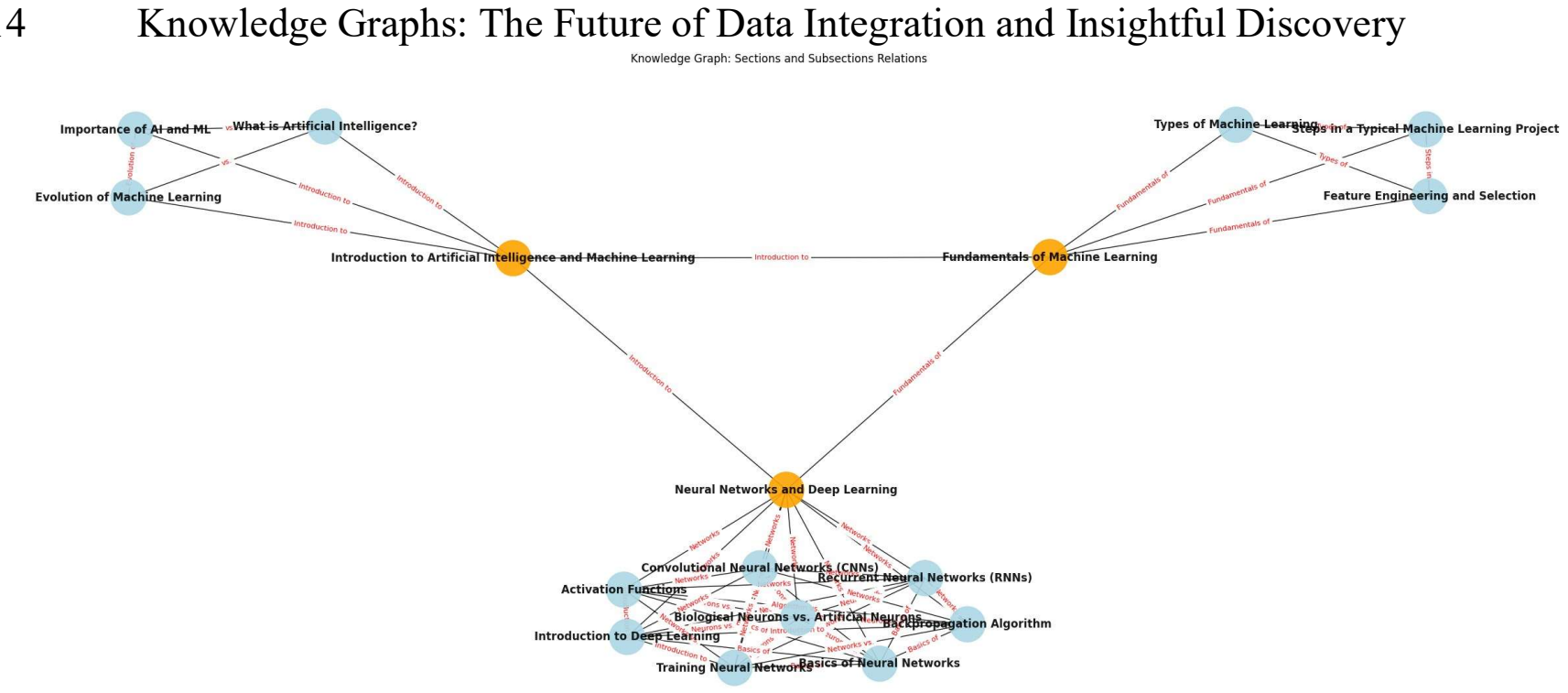

Fig. 6. The knowledge graph with additional predicting relations without using get-entities .

**RDF clustering**

the RDF data was loaded into an RDFLIB graph representing the same nodes of entities connected with relations. each node was mapped to a numeric value, and the embeddings in Fig 7 were extracted using the DeepWalk model from the KarateClub library. The elbow method in Fig 8 found the optimal number of clusters for applying the K-means. Based on the embeddings. Finally, the accuracy of the clustering was measured by measuring the similarity between the nodes using the silhouette score.

**K-means clustering**

After the knowledge graph was built by extracting the entities and relations from some Wikipedia sentences, the node2vec algorithm was applied, and also the DeepWalk model was tried to generate the node embeddings for the graph, and then the embeddings were in 2-dimensional as in Fig 10. The number of clusters was chosen based on the silhouette score. As a loop was applied on different values of the number of clusters, the final number should be the best number of clusters for this data based on the embeddings that catch the large node similarity and the best number of clusters. This number of clusters was applied to the K-means, and the nodes were visualized with different colors in Fig 9 based on the cluster assigned to this specific entity.

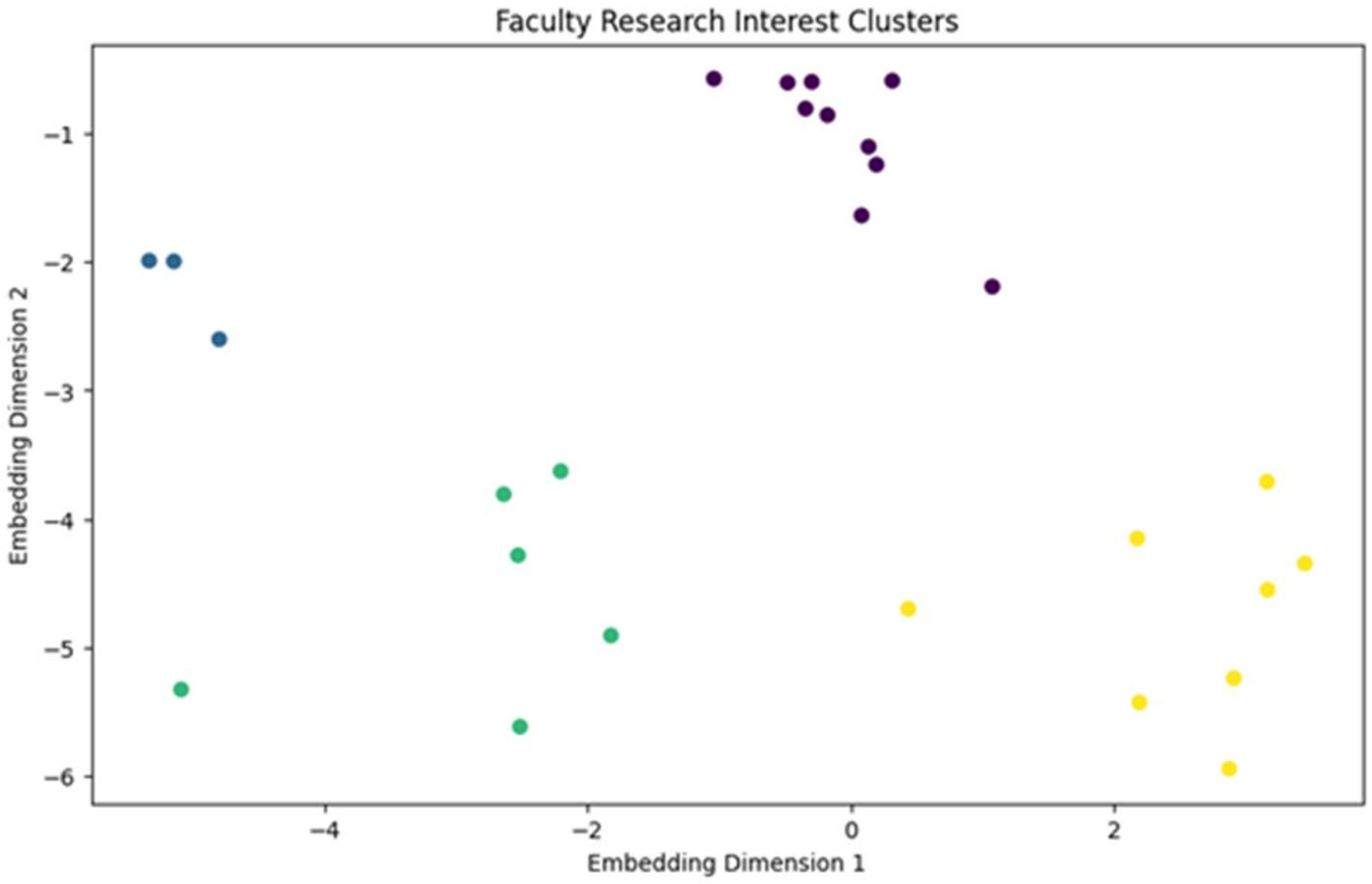

Fig. 7. The Embedding Clusters.

**Gap statistics**

the gap statistics were computed to determine the optimal number of clusters for using them in k-means clustering as in Fig 11, but it proved the lowest accuracy of similarity between the entities in the cluster.

**Agglomerative clustering**

The knowledge graph was built on the same data as Wikipedia Sentences, and the graphs' embeddings were extracted using the node2vec model and the Deep-Walk model as presented in Fig 12. The optimal number of clusters was chosen using the silhouette score after iterating on the maximum number of clusters to indicate which represented the best number of clusters. Then according to Fig 13 the nodes were clustered using agglomerative clustering.

After applying the clustering method, LLama 7b after fine-tuning it on our Wikipedia dataset, and then, the fine-tuned model generated a cluster descrip-tion that describes the relationship or the point of similarity that was applied between the entities, as for an example prompt was ”Give me one specific descrip-tion that describes the similarity between the nodes in the cluster: node$_l$abels.”. Different models were used for training it on the dataset, like GPT2 and Falcon-7b, but LLama provided the most accurate results.

These clustering methods have limitations in that after each run; some enti-ties can leave and go to another cluster.

**ExCut: Explainable Clustering**

To address the limitations of traditional clustering methods, such as the insta-bility of cluster assignments and the lack of interpretability, ExCut Gad-Elrab et al. (2020) is designed to compute explainable entity clusters over knowledge graphs. The central challenge it tackles is the clustering of a vast array of entities from a knowledge graph into meaningful groups. However, merely clustering the entities does not sufice; the clusters must also be explainable and interpretable to users. They necessitate human-comprehensible labels that succinctly describe the essence of each cluster.

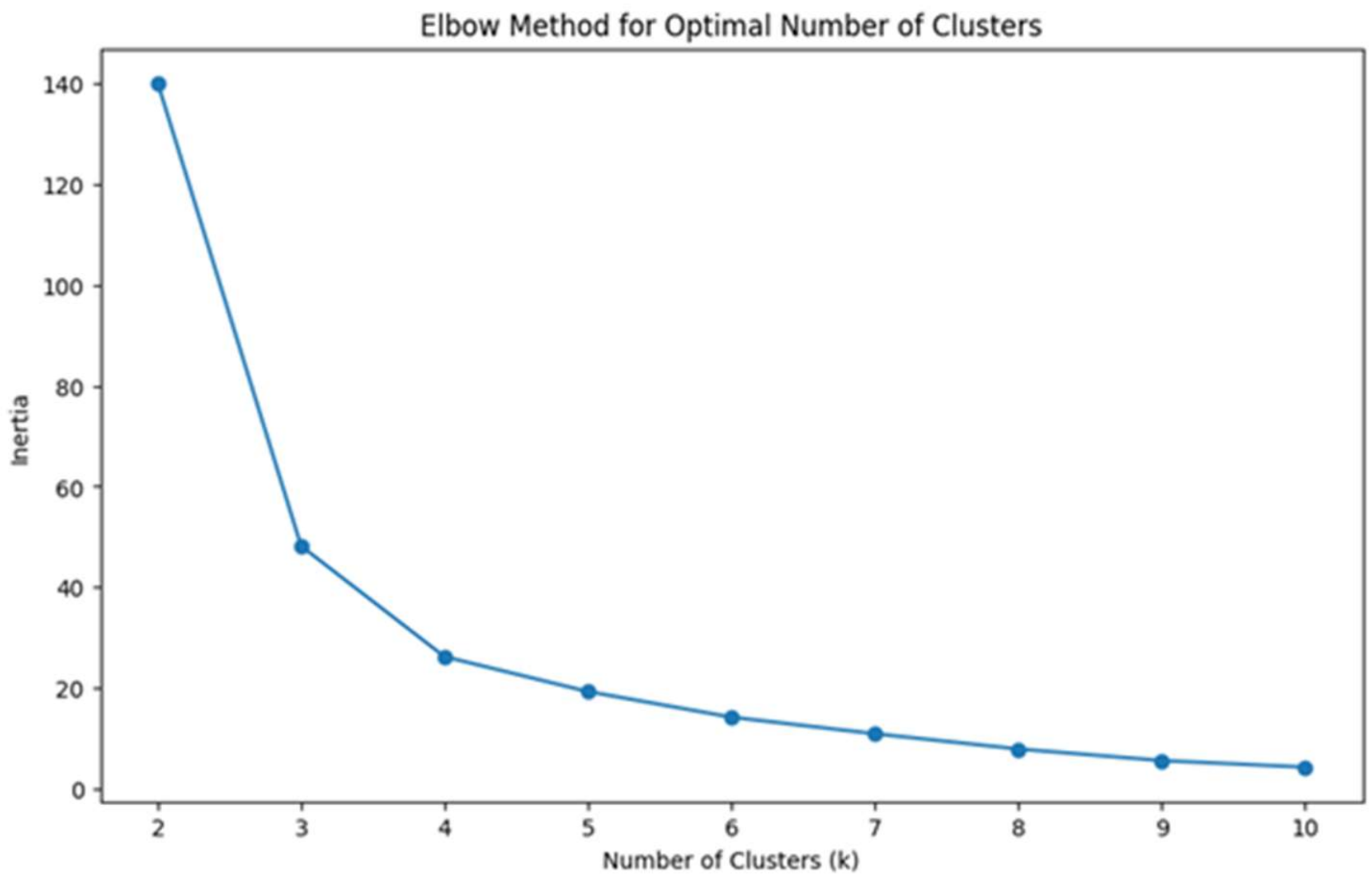

Fig. 8. The Elbow Graph.

Table 2. Accuracy of the clustering method with the change of the embeddings ex-traction model

| Model | K-means | Agglomerative clustering |
|---|---|---|
| **Deepwalk** | 0.68 | 0.53 |
| **Node2vec** | 0.63 | 0.64 |

ExCut adopts a joint approach that encompasses both clustering and expla-nation mining. Initially, it learns embeddings for entities using models such as TransE and ComplEx. Subsequently, it performs initial clustering of target en-tities based on their embeddings. Concurrently, ExCut learns explanation rules for each cluster directly from the knowledge graph. These rules are then applied to infer new entity-cluster assignments. Importantly, the feedback from these assignments is used to fine-tune the embeddings, thereby guiding the clustering towards more coherent and explainable groups. The process iterates between clustering, rule learning, and embedding adaptation to enhance both the quality of the clusters and the clarity of their explanations.

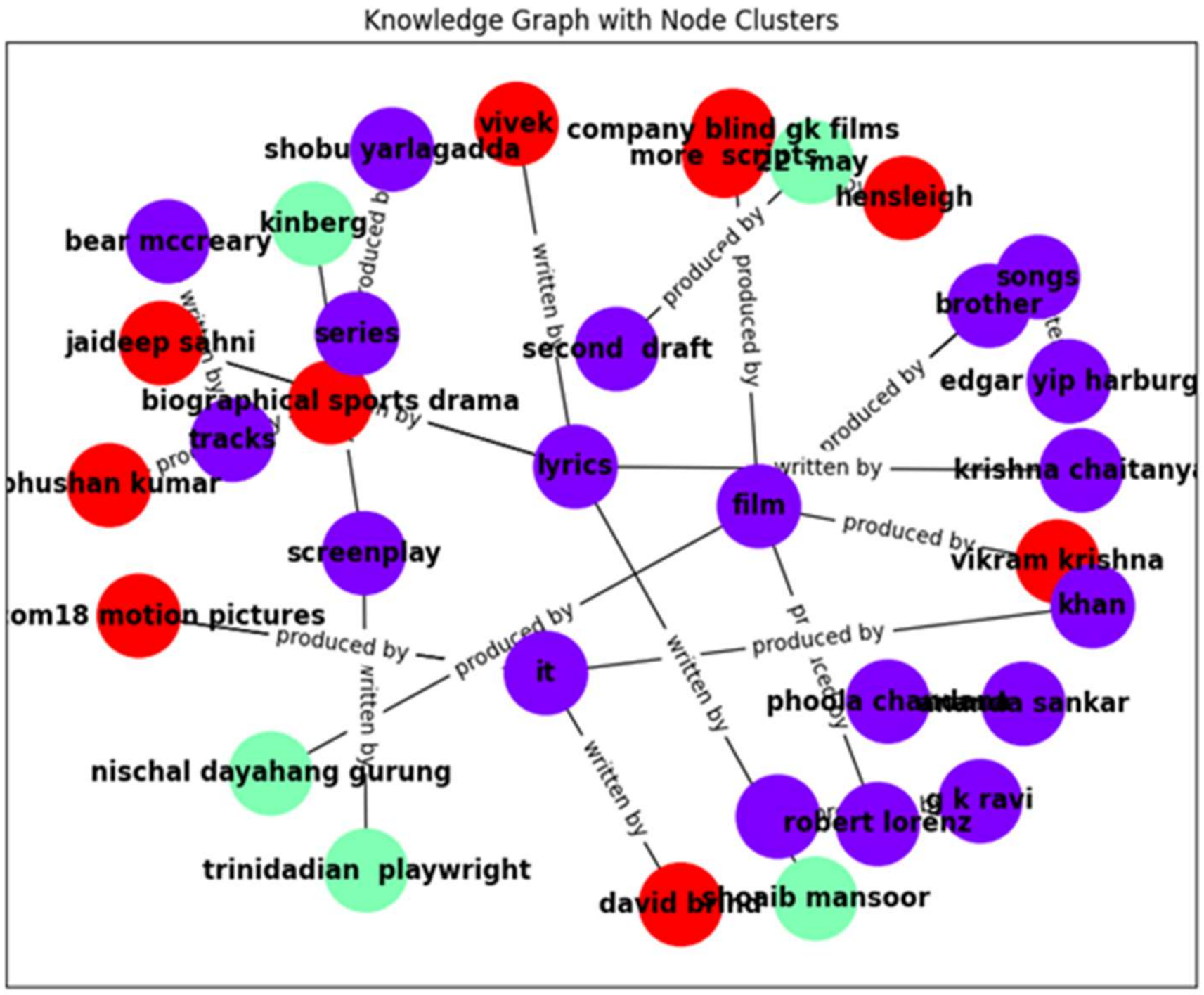

Fig. 9. The Clusters of the Nodes Using K-Means .

Experiments have demonstrated that ExCut generates clusters of higher quality and greater interpretability compared to baseline methods on real-world knowledge graphs. In essence, ExCut adeptly addresses the problem of comput-ing not just any entity clusters but specifically explainable clusters over knowl-edge graphs, by synergistically combining embedding-based clustering with rule learning in a dynamic, iterative process Gad-Elrab et al. (2020).

## 6.2 Graph Neural Network in Link Prediction and Adding New En-tities

One of the major challenges faced with knowledge graphs is disconnected sub-graphs, where entities are not fully connected within the graph. When subgraphs are disconnected, the entities contained within them cannot benefit from poten-tial relationships and links to other parts of the knowledge graph. This fragmen-tation of information limits the insights that can be discovered from the graph. One effective approach for addressing disconnected subgraphs is link prediction using graph neural networks (GNNs). GNNs have the advantage of being able to learn representations of nodes by aggregating features from their neighbor-ing nodes in the graph. RAGAT Liu et al. (2021) is a specialized GNN model designed specifically for the task of link prediction within knowledge graphs. RAGAT predicts a score between 0 and 1 for each pair of entities, with scores closer to 1 indicating higher confidence that a link exists between the two enti-ties. RAGAT has an attention mechanism gives it several advantages over other link prediction models. Unlike methods that iterate over each node in the graph, RAGAT can identify the most nodes for predicting specific links. It assigns at-tention weights to nodes based on their predicted importance to the link. This allows RAGAT to focus its computations only on the subset of nodes most rel-evant to the prediction task. As a result, RAGAT can perform link prediction much more eficiently on large graphs compared to approaches that consider all nodes equally. The attention mechanism enables RAGAT to scale effectively to industry-sized knowledge graphs with millions or billions of entities and relation-ships. Its ability to quickly identify missing links helps address the fragmenta-tion caused by disconnected subgraphs. RAGAT has shown promising results in evaluating its ability to predict missing links. RAGAT was tested on two stan-dard knowledge graph link prediction benchmarks. The datasets are divided into three parts: test, validation, and training as seen in table 3. Each part follows the format of entity, relation, and entity. On the WN18RR benchmark dataset, RAGAT achieved a Mean Reciprocal Rank (MRR) score of 0.489 and a hit@10 score of 0.562. MRR is a common metric for evaluating link prediction mod-els, where higher scores indicate better performance. An MRR of 0.489 means that on average, the correct missing link was ranked in the top two predicted links by RAGAT. RAGAT was also evaluated on the FB15K-237 benchmark dataset, achieving an MRR of 0.365 and hit@10 of 0.547 as shown in table 4. These scores demonstrate RAGAT's effectiveness at the task compared to other GNN and baseline models on both datasets.Its attention mechanism allows it to identify high-confidence missing link predictions to help reduce fragmentation in knowledge graphs. The results on the three standard benchmarks validated RAGAT's ability to predict missing links with competitive performance.

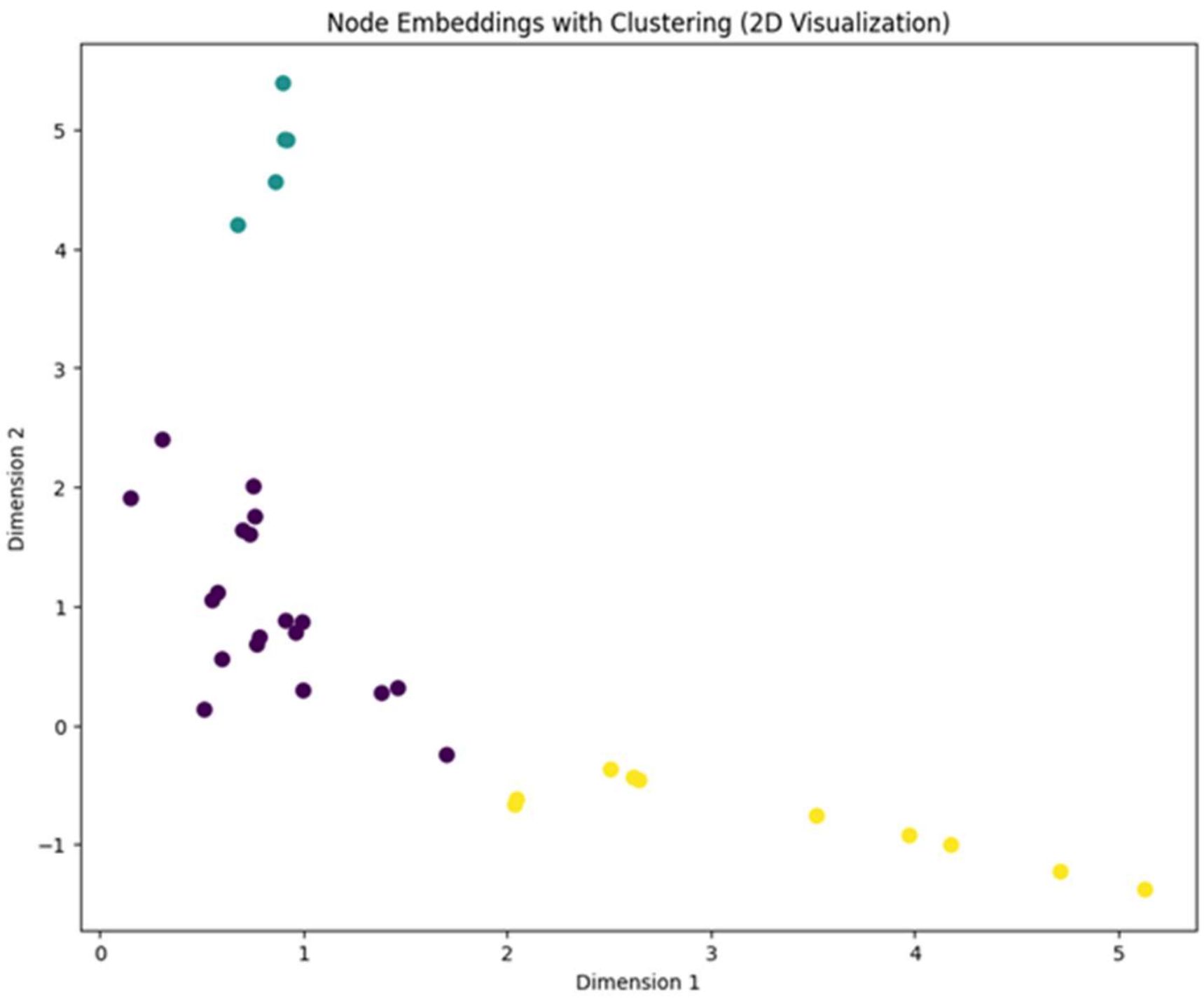

Fig. 10. The Embedding Clusters Using K-Means.

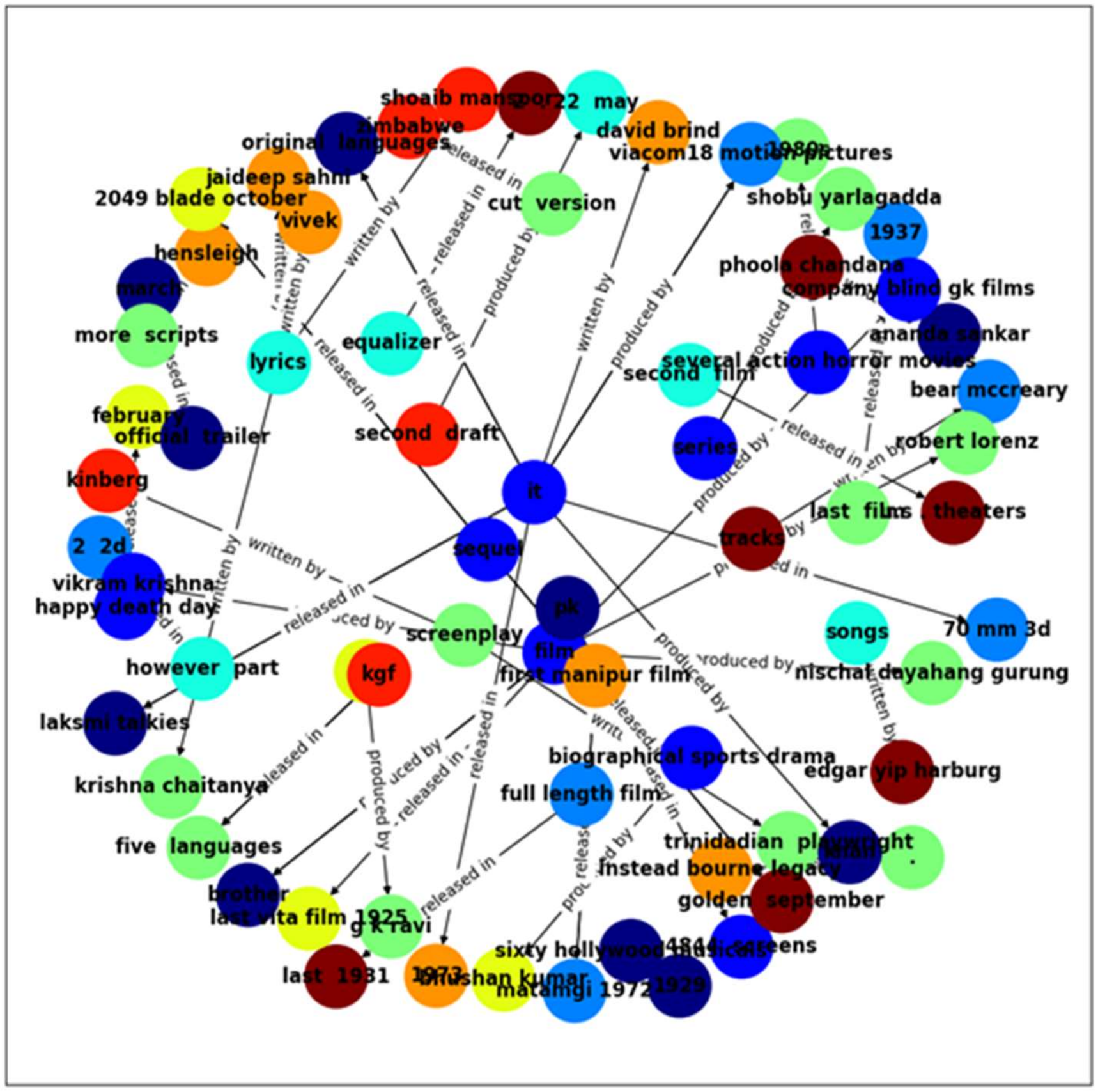

Fig. 11. The clusters using gap statistics .

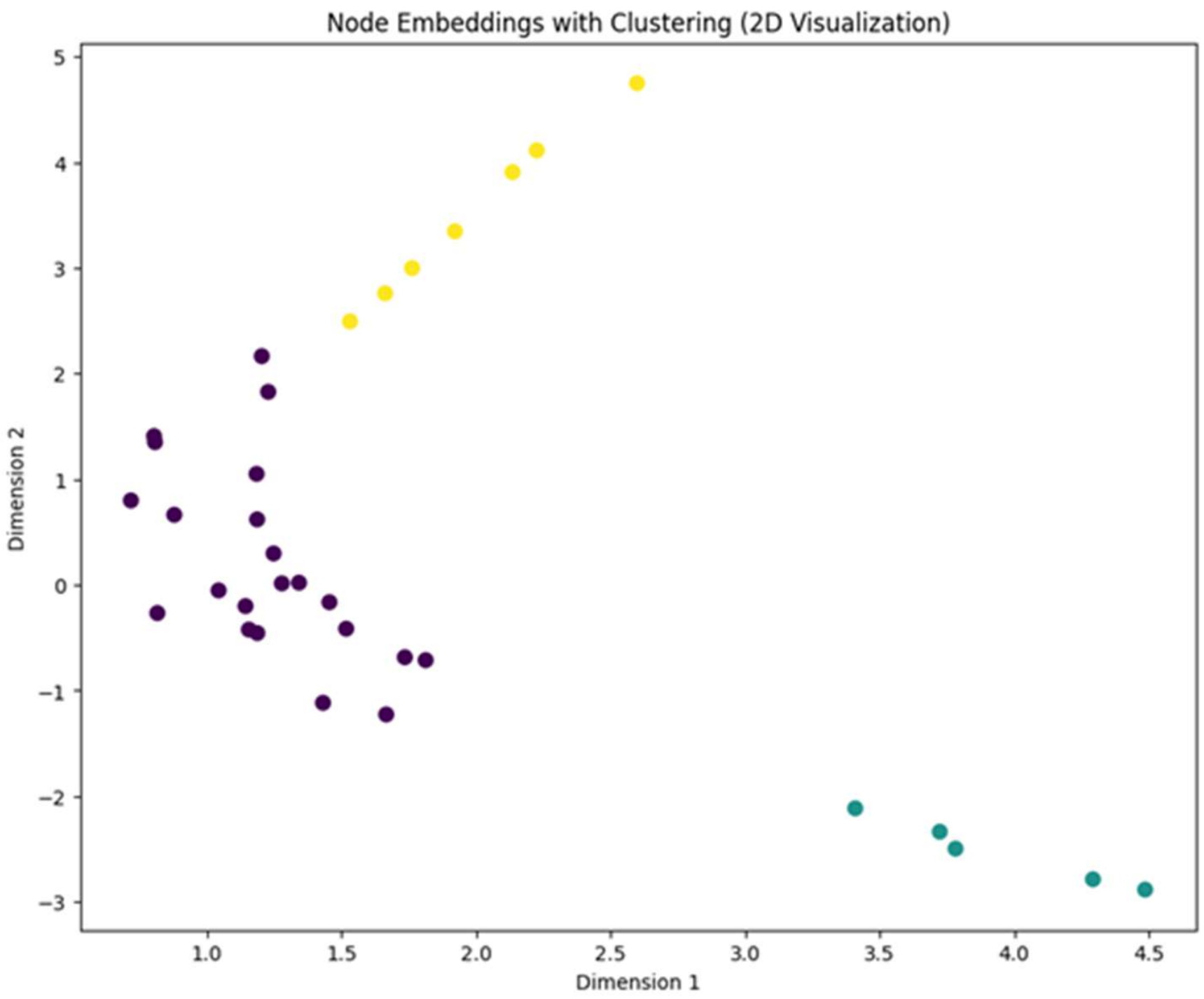

Fig. 12. The embeddings clusters using agglomerative clustering . Table 3. Dataset Statistics

| Dataset | Train | Validation | Test |
|---|---|---|---|
| **FB15k-237** | 272,115 | 17,535 | 20,466 |
| **WN18RR** | 86,835 | 3,034 | 3,134 |

Table 4. RAGAT Performance on Different Benchmark Datasets

| Dataset | MRR | Hit@10 |
|---|---|---|
| **WN18RR** | 0.489 | 0.562 |
| **FB15K-237** | 0.365 | 0.547 |

### 6.3 Traditional Embedding Techniques in Link Predictions and Adding New Entities

Traditional embedding techniques like TransE learn embeddings for entities and relations in knowledge graphs to predict missing links. These models learn low-dimensional vector representations of entities and relationships where similar entities and relationships are closer in the embedding space. The models are rel-atively simple and eficient to train. However, these techniques also have several

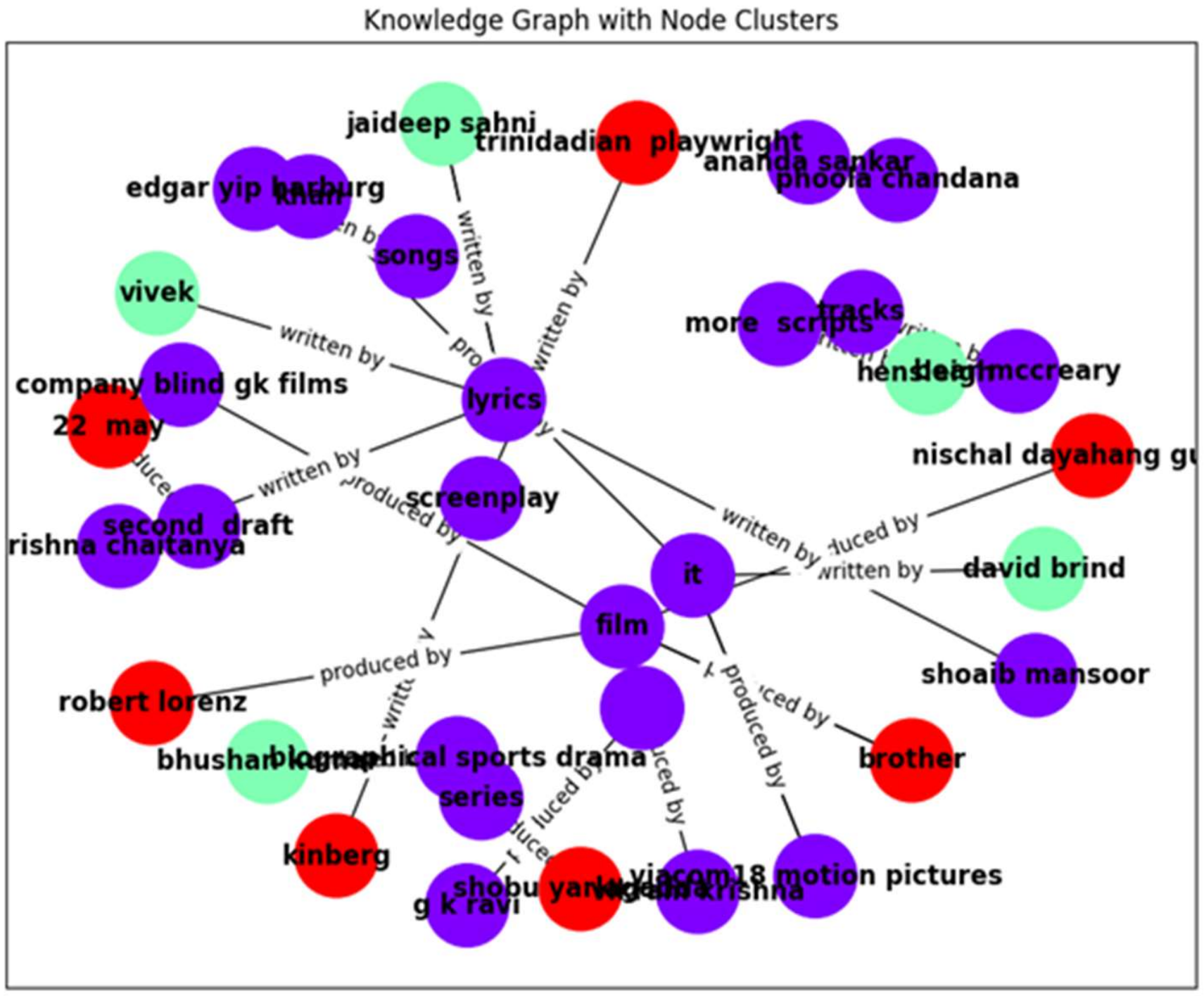

Fig. 13. The clusters of the Nodes using agglomerative clustering .

limitations. They struggle to incorporate new entities, as adding a new entity requires retraining the entire model from scratch, making them less suitable for dynamic knowledge graphs. The models also treat all relations independently without capturing correlations between different relation types, limiting their ability to make accurate predictions involving multiple relations. The learned embeddings may not generalize well to new entities and relations not seen dur-ing training. Furthermore, traditional techniques like TransE assume a single relation between any pair of entities, which may not reflect real-world knowl-edge graphs that can contain multiple relations. More recent techniques aim to address some of the limitations of traditional embeddings.
TransE Model

The “TransEModel” class is a pytorch neural network module. The model would take entity indices for the head, relation, and tail as input entities to compute the distance between the head, relation, and tail embeddings using the L1 norm. The trans model was trained using SGD (stochastic gradient descent), and the marking ranking loss was used to know the positive and negative samples. The negative sample was generated by corrupting the positive samples, and this idea is to create a false statement that doesn't hold in KG, so the model noticed the difference between the positive and negative samples.

The relation was predicted between two entities using the transE model by computing the distance for all the relations and returning the relation index and distances. All the relations were tested between each pair of entities to find the strongest relation with the smallest distance. The relation refers to the numerical value for representing similarity and dissimilarity. As in the transE model, the goal is to learn embeddings for entities and relations so that the distance between the embeddings reflects the semantics of relationships in the knowledge graph. For example, There was some entities of the knowledge graph that were connected to each other, so if the middle node was deleted, the mission is to predict the relation between the two other entities along the distance. If the three nodes were B, A, and C, the distance between B and C differed from C to B. So, the smallest distance with the strongest relation should be chosen according to Fig 14.

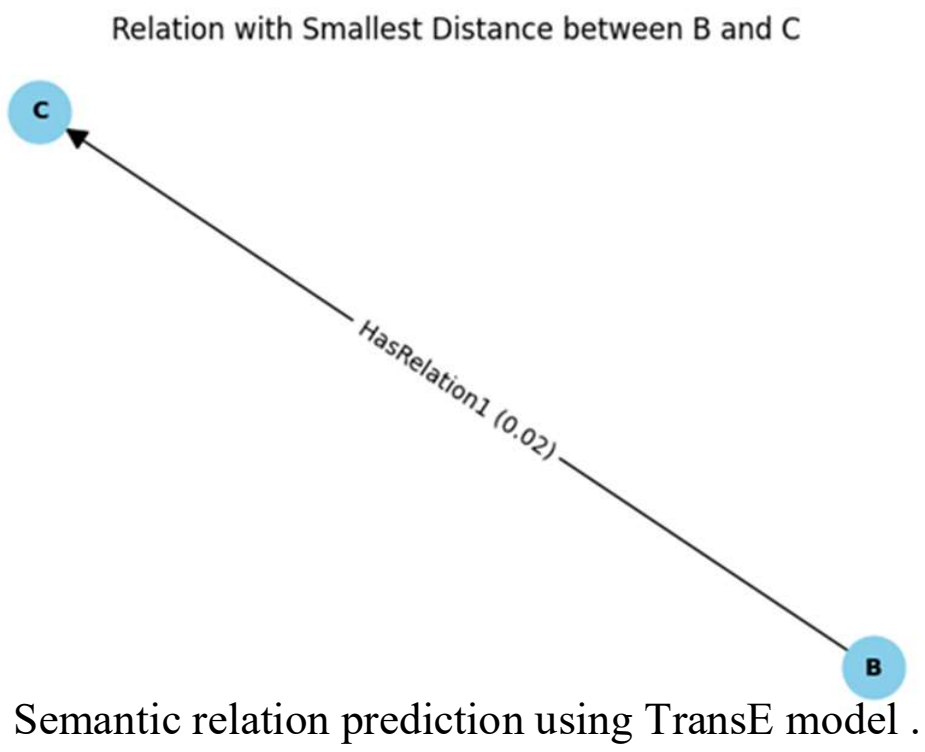

Fig. 14. Semantic relation prediction using TransE model .

### 6.4 Extract triples and predict relations using pre-trained ontology

there are three distinct methods for extracting triples and predicting relations between entities: DBpediaOntology class connected to the API of DBpedia Spot-light, ConceptNetOntology connecting with an API from ConceptNet, and on top of that, Rebel model retrieved from Babelscape assemblage. Every approach targets the problem of information retrieval from unstructured text; however, they vary in the applied approaches and purposes.

**DBpediaOntology**

The DBpediaOntology class uses the DBpedia Spotlight API to extract triple from a text. This process first involves setting a confidence threshold for the API annotations and how this system filters through them. The class then forms the parameters for an API request, including input text and confidence threshold, before issuing a call to the REST endpoint. The API response is then processed to get important information about entities mentioned within the document, in-cluding Uniform Resource Identifiers and surface forms. These extracted triples consisting of subjects, predicates, and objects represent relationships in the in-put text in a structured way. As a connection to the data in DBpedia knowledge base, it enriches information extracted from text by introducing connections with an enormous database. This method is especially useful in cases where a clear and valid knowledge base linkage implies establishing reliable access to detailed data on entities mentioned within the target text, making further analysis or integration with external sources available.

**ConceptNetOntology**

The ConceptNetOntology class interacts via the concept's API to predict rela-tions between two Input concepts. The class takes the input concepts and builds API parameters around them, which then sends a request to ConceptNet's API service that returns relations regarding relevant edges from concept relation-ships with relation labels. This method emphasizes the construction linking of concepts grounded in the ConceptNet knowledge base, which tries to reveal se-mantic structures between entities.

**REBEL**

In the process of converting unstructured text to triples and predicting relations, Rebel Huguet Cabot and Navigli (2021) uses a machine-learning approach. The main function is "extract-relations-from-model output," which uses the examples of text by tokenizing and finding tokens that represent relations like subj and obj. These tokens direct the extraction of structured triples from model predic-tions in the form of subject, relation type, and object. The approach systemati-cally accumulates these triples as it walks through the decoded output, utilizing patterns learned from and context information in the pre-trained Rebel model. The model automatically organizes data extracted from the document by de-tecting long-range dependencies and semantic relations. Such relation-extracting methodology, driven by machine learning herewith, provides a framework that is easily configurable and adaptable in situations where Rebel code can extract insights and information as long as it was tested on some triples that were on the same pre-trained domain(e.g., Wikipedia Data). The Rebel model reveals the development of natural language understanding, which can be implemented for relation extraction from heterogeneous text data.

These methods had many problems with unknown triples belonging to these on-tologies, as they trained on some triples in the form of (subject, relation, object). If one of these entities were changed, then the model wouldn't get any relation predicted.

# 7 How can retrieved information from knowledge graphs play an important role in different domains? (RQ3)

## 7.1 Knowledge graph in Explainable AI

KG explainability can be utilized at different stages in the AI development pro-cess. Typically, KG explainability is performed before (pre-modeling explainabil-ity), during (explainable modeling), or after (post-modeling explainability) the AI modeling stage Rajabi and Etminani (2022).

Pre-modeling Explainability: This method works independently of the model and usually involves using a KG before selecting a model, as it applies only to the data itself. Pre-modeling explainability can include constructing KGs from a dataset or standardizing a dataset with KGs. For instance, in one study, a KG was used to transfer features in a zero-shot learning model and generate explanations for unseen classes in an image classification problem Geng et al. (2021).

In-modeling Explainability: This approach focuses on the model's inner workings, such as its mathematical aspects, and uses KGs to generate expla-nations during the training phase. For example, a KG was used to improve a deep learning model's performance on an image classification task by training the model to predict every node in a knowledge graph and then propagating information between the nodes to refine predictions Daniels et al. (2020).

Post-modeling Explainability: These techniques describe the application of a KG after a model has been trained. They enhance the explainability of AI models by using KGs to provide insights into what the trained model has learned without altering the underlying model. For example, in one study, a KG was applied to a graph neural network model to capture important features and relationships in a set of news articles, using the relevance scores of entities to guide the embedding of the articles Cui et al. (2020).

## 7.2 Knowledge Graph for Automatic Coding

Knowledge graphs have been proven extremely useful in powering diverse ap-plications in various applications such as program search, code understanding, bug detection,and code automation. GraphGen4Code Abdelaziz et al. (2021) is therefore designed as a toolkit to build knowledge graphs for program code. This toolkit can construct code representations that represent actual program flow along with natural language descriptions of API calls when they exist to enhance code representations.

COCOMIC Ding et al. (2022) is a framework on top of existing code lan-guage models that jointly learns in-file and cross-file context for code comple-tion.Authors also build CCFINDER. The tool has two main steps: (1) Analyze the program dependencies within the project and parse the source code to ex-tract both the bird's-eye view of the whole project and the code details of each module. With that,CCFINDER builds the project context graph: graph nodes represent code components that constitute the project's backbone,and edges in-dicate the relations among these components. (2) Given an incomplete program, the tool will identify the most relevant cross-file context and retrieve their details from the built graph.

### 7.3 Knowledge Graph for Self Driving Car

Zheng and Kordjamshidi (2022) propose the Dynamic Relevance Graph Network (DRGN) model to tackle the challenging problem of commonsense question an-swering using external knowledge from a knowledge graph. Existing approaches have dificulty reasoning over knowledge graphs when edges are missing in the extracted subgraph needed for reasoning chains, or when handling questions with negative words. DRGN operates on a knowledge graph subgraph contain-ing entities mentioned in the question and answers. It adds the question as a node to provide contextual information . Unlike prior models, DRGN com-putes dynamic relevance matrices between node representations at each layer of a relational graph network. This allows it to establish direct connections and potentially recover missing edges by learning new edges based on chang-ing relevance scores. Computing relevance between the question node and entity nodes also helps leverage contextual cues better [1]. Experiments on the Com-monsenseQA and OpenbookQA benchmarks demonstrate that DRGN achieves state-of-the-art performance, obtaining particularly strong results on negatively worded questions through its use of question-entity relevance. Ablation studies validate that component such as the knowledge graph, relational edges, question node, and dynamic relevance computation contribute to DRGN's effectiveness at capturing relationships for improved commonsense question answering.

Krause (2022) proposes the Navi Approach as a method to enable dynamic knowledge graph embeddings. Currently, most existing embedding methods as-sume a static knowledge graph structure. When updates occur through additions or deletions of nodes and edges over time, the embeddings need to be retrained from scratch. However, retraining for large graphs is computationally expensive and can affect downstream machine learning models that utilize the embeddings. The Navi Approach works to address this issue by reconstructing embeddings locally based only on the neighborhood of each node, rather than requiring a full retraining of the embedding model. It leverages existing static embeddings as a starting point and independently derives new embeddings for nodes based on the embeddings of neighboring nodes. Two different types of ”Navi Layers” are presented - the Connectivity Layer which takes the average of neighbor embed-dings, and the Translational Layer which applies transformations like TransE. Composite reconstructions can be obtained by combining multiple Navi Lay-ers . Preliminary evaluation on benchmark datasets shows the reconstructed embeddings with the Navi Approach can maintain or outperform the accuracy of retraining embeddings from scratch for entity classification, indicating the hypotheses around enabling dynamic embeddings may be answered positively. The approach is evaluated in tasks like link prediction and entity classification with simulated graph updates, comparing different methods for handling up-dates. Initial results are promising in maintaining state-of-the-art performance without requiring full retraining of embeddings.

Hogan et al. (2020) introduces a tool called the Dynamic Knowledge Graph (DKG) that aims to facilitate conceptual change in knowledge building environ-ments. DKG analyzes student notes from Knowledge Forum to automatically generate dynamic knowledge graphs that visualize core concepts and relation-ships within a domain. It does this through three main phases - extracting in-structional concepts from notes using techniques like POS tagging and TF-IDF, identifying relationships between concepts using association rule mining, and displaying interactive knowledge graphs. A preliminary study found that the central concepts identified in a physics knowledge graph aligned with the rele-vant course content. However, the paper discusses limitations with the current system and areas for improvement, such as integrating it more seamlessly with Knowledge Forum and allowing for manual edits. The goal of DKG is to support collaborative conceptual change over time by making conceptual insights from student work more explicit and dynamic through automated knowledge graph generation and analysis .

Das et al. (2018) proposes a new neural machine reading model called KG-MRC that can build dynamic knowledge graphs from procedural text to track how entity states change over time. KG-MRC constructs a knowledge graph for each step in a described process by updating the graph from the previous step. It uses a machine reading comprehension module to query the current state of entities by asking questions like ”Where is the entity located?” . The MRC module conditions on both the text and the knowledge graph from the previous step. KG-MRC also employs soft co-reference mechanisms to handle different text mentions of the same entity or state over time. It was evaluated on two tasks from the PROPARA dataset about tracking entity state changes in processes, outperforming previous models. KG-MRC also achieved state-of-the-art results on predicting ingredient locations from recipes in the RECIPES dataset. The knowledge graphs built by KG-MRC at each step help improve its understanding of the text for downstream question answering tasks.

Lavdim et al. (2020) presented that AD datasets capture sensor data but lack semantic relationships between entities, this problem can be solved some-how using Knowledge graphs can represent richer semantics and relationships for driving scenes, which enhance deep learning techniques used for AD tasks. Stan-dardized ontology also can be used to enable integrating multiple datasets, where an architecture, ontologies for AD datasets, and knowledge graphs are present in the structure which was a 3-layer architecture with raw data, knowledge graphs, and applications. Ontologies for nuScenes, BDD100K, and PandaSet datasets based on their structure and concepts were created, built knowledge graphs by converting annotation data to RDF triples based on the ontologies, nuScenes KG has 40K scenes, BDD100K has 100K scenes, PandaSet has 8K scenes, and generated a global KG by merging common concepts from the dataset KGs. The improvements in the eficiency after the process were: The knowledge graphs cap-ture richer semantics like objects, relationships, and scene context. Outline AD tasks that could benefit from KGs: object detection, segmentation, prediction, and planning. KGs enable new applications like multi-dataset training, domain adaptation, and scene retrieval. Provide sample SPARQL queries to retrieve use-ful information from KGs. KGs can provide contextual knowledge to augment AD pipelines.

Gad-Elrab et al. (2020) introduces a novel methodology termed ExCut, which is designed to compute explainable entity clusters over knowledge graphs. The central challenge it tackles is the clustering of a vast array of entities from a knowledge graph into meaningful groups. However, merely clustering the enti-ties does not sufice; the clusters must also be explainable and interpretable to users. They necessitate humancomprehensible labels that succinctly describe the essence of each cluster . To address this, ExCut adopts a joint approach that en-compasses both clustering and explanation mining. Initially, it learns embeddings for entities using models such as TransE/ComplEx. Subsequently, it performs initial clustering of target entities based on their embeddings. Concurrently, Ex-Cut learns explanation rules for each cluster directly from the knowledge graph. These rules are then applied to infer new entity-cluster assignments. Impor-tantly, the feedback from these assignments is used to fine-tune the embeddings, thereby guiding the clustering towards more coherent and explainable groups. The process iterates between clustering, rule learning, and embedding adaptation to enhance both the quality of the clusters and the clarity of their explanations. Experiments have demonstrated that ExCut generates clusters of higher quality and greater interpretability compared to baseline methods on real-world knowl-edge graphs. In essence, ExCut adeptly addresses the problem of computing not just any entity clusters but specifically explainable clusters over knowledge graphs, by synergistically combining embedding-based clustering with rule learn-ing in a dynamic, iterative process.

Olszewska et al. (2017) presented the development of knowledge graph ontol-ogy and reasoning in autonomous robotics. Ontology provides common semantics for concepts used in autonomous robot systems, enabling clear communication between robots and humans. The paper validates this in a human-robot interac-tion scenario. The ontology in this paper builds on existing ontologies in SUMO (Suggested Upper Merged Ontology), which organizes the definition of domain-specific ontologies. These define various physical concepts, processes, quantities, and relations. These concepts are axiomatized in the first order and modeled in OWL. CORA (Core Ontology for Robotics and Automation) extends SUMO with its robotics-specific concepts like robots, robot motion, and robot behavior . CORA defines a device taxonomy with sensing, processing, actuating, and com-municating devices, then model's robot components, capabilities, and environ-ments where robots operate. The ontology in this paper is implemented in OWL (Web Ontology Language) as OWL enables the definition of classes, properties, individuals, and axioms. It supports reasoning and inferencing to derive new knowledge. In summary, SUMO provides high-level concepts, CORA extends those to robotics, and OWL provides the language to implement and reason with the ontologies. The ontologies were validated in a human-robot interaction scenario with multiple robots and humans. Concepts like ”hasCommunication-With” allow communication links between humans and robots automatically. Integrated with the core ontology, the visual ontology enabled processing vision data from robot scenarios to extract information like objects or scenes. Rea-soning on the ontology was done using an off-the-shelf reasoner engine. Overall, the ontology demonstrated strengths in enabling semantic interoperability and relating different ontologies for knowledge sharing.

Bosch et al. (2021) presented a survey on knowledge graphs (KGs) applied to automated driving (AD). KGs can represent complex structured and rela-tional knowledge, addressing limitations of deep learning methods for AD, such as verifiability, explainability, and safety. The paper reviews ontologies and KG-based approaches, categorizing them based on AD components like perception, scene understanding, and motion planning. The ontology of the paper presents autonomous concepts covering the route, vehicle, driver, and context, for ex-ample. KG-based approaches are analyzed for object detection, mapping, scene understanding, behavior prediction, motion planning, and validation. These ap-proaches are classified based on the AD components and sub-tasks they ad-dress. The KG defines an ontology that models key concepts and their relations. Domain-specific ontologies are created to capture automotive knowledge. The knowledge graph is populated with instances of real-world entities and their re-lations based on the ontology schema. Knowledge can be extracted from sensor data, maps, trafic rules, and driver manuals. The decision-making module typ-ically utilizes a reasoner or inference engine to query the knowledge graph and derive new facts based on the encoded logic. The correctness of decisions can be validated by checking their consistency with the knowledge graph facts and con-straints. Explanations can be generated from the knowledge graph to increase trust and transparency in decisions. Ultimately, performance must be validated on real-world driving data across diverse scenarios. KGs have been applied for perception, scene graphs, lane detection, segmentation, and mapping. For mo-tion planning, KGs help encode rules and driving knowledge for decision-making. KGs also validate through risk assessment, test generation, and requirements ver-ification [8]. However, only a few KGs exist for each AD component, suggesting a need for future research.

Tan et al. (2021) constructed a KG in the realm of urban trafic and predicted the relationships between trafic entities. The paper highlights the importance of KG reasoning models with the TransD model. It plays a significant role in congestion detection, with applications in knowledge discovery and intelligent query answering. It also aids in the discovery of similar travelers and the ability to find the shortest path between trafic intersections. The study adopts a top-down approach to construct a KG for urban trafic. Knowledge is constructed, and the storage of entities, their attributes, and the relationships between these entities are managed using a graph database and Neo4j. The paper introduces the reasoning KG model, which transforms the semantic information of enti-ties and relationships into low-dimensional vectors. This enables calculations for tasks like entity linking and relationship reasoning. The model employs the TransD model, which uses two vectors to represent entities and relationships. A mapping matrix is constructed by a projection matrix. These vectors, cal-culated in the relational space, are measured using a score function based on the L2 Euclidean distance. The authors also use L2 norm constraints to opti-mize model parameters. The study proposes a method for constructing negative samples and employs the mini-batch gradient descent approach during model training. The authors evaluate the TransD model by performing link prediction tasks. The evaluation metrics are MRR (Mean Reciprocal Rank) and average rank (MR). Predicting both head and tail entities in the knowledge reasoning model indicates that the TransD model is effective, with an MRR of 0.35, MR of 33 for head entity prediction, and an MRR of 0.23 and MR of 37.5 for tail entity prediction. Thus, the model accurately predicted the correct entity compared to optimal results.

Halilaj et al. (2021) introduced a knowledge graph and a graph-based neu-ral network for enhancing situation understanding in automated driving systems. The authors offer the CoSI (Context and Situation Intelligence) approach, which leverages knowledge graphs to represent and comprehend complex driving sce-narios. This study presents the CoSI knowledge graph (CKG), which is a seman-tic representation of the driving environment, capturing structured knowledge from heterogeneous sources. It is designed with principles that support easy extension and exploration, allowing for the incorporation of various ontology concepts relevant to driving scenarios such as scenes, situations, scenarios, obser-vations, drivers, and user preferences. CKG is populated with data transformed and enriched from sensor data, providing an RDF representation of real-world information. Both declarative and imperative approaches are employed for this purpose, enabling the mapping of sensor data to ontological concepts. The CKG contains an extensive set of classes, object datatype properties, and annotation properties, forming a rich and structured knowledge base capable of representing complex driving scenarios. These are essential for enabling enhanced situation comprehension in the context of highly automated driving systems. The authors focus on classification learning and prediction. The performance of the KG-based approach was compared to vector-based algorithms. They used various machine learning algorithms, including the support vector machine, the decision-tree clas-sifier, and the Multi-Layer Perceptron (MLP). The results of the classification demonstrate the effectiveness of the KG. The authors used the five most impor-tant features, achieving over 95 accuracy in classifying driving situations. The results indicate that the knowledge graph provides more discriminative informa-tion than traditional feature vector-based representation, especially in complex situations involving multiple relationships and participant interactions. The KG performs well in complex tasks that consider a rich context.

Zhao et al. (2016) introduced the central idea of enabling an autonomous vehicle to comprehend complex trafic situations and make real-time decisions. They aim to improve the decision-making process for autonomous vehicles by using the ontology of sensor data in a machine-understandable format. The on-tology consists of concepts (classes) and relationships (properties). Instances are described using Resource Description Framework (RDF) triples as ¡subject, property, object¿. The Semantic Web Rule Language (SWRL) expresses traf-fic rules, while SPARQL is employed for querying RDF data. This knowledge base is essential for the decision-making system to understand and process data from the vehicle's sensors [11]. The paper presents the SPARQL query engine, which retrieves information from the KG. A sub-KG is created based on the current position of the vehicles and contains only relevant information to reduce decision-making time. The system identifies road segments directly connected to the vehicle's current position using SPARQL queries. The system extends the search to road segments within a 2-depth range. This is important for consid-ering the broader context, for example, if the road has two lanes. The sub-KG also contains SWRL rules essential for inferring driving decisions. The SWRL rule reasoner decides when a collision warning is received. It uses the sub-KG to make driving decisions based on the received data and predefined SWRL rules. The system uses the Pellet and OWL API to reason with SWRL rules, allowing it to infer safe driving decisions promptly. The research conducted real-world experiments using a TOYOTA ESTIMA equipped with various sensors, such as Velodyne Lidar, GPS-IMU, and cameras. The experiment included uncontrolled intersections and narrow roads. However, decisions were made correctly, and the sub-KG reduced the decision-making time, with an average time of 53ms, making it close to the duration of sensor data transmission.

Wickramarachchi et al. (2020) Managing this data eficiently is crucial for de-veloping and deploying autonomous vehicles.Explored knowledge graphs (KGs) as a potential solution to this data management challenge. Their research, titled "An Evaluation of Knowledge Graph Embeddings for Autonomous Driving Data: Experience and Practice," delves deep into generating and evaluating knowledge graph embeddings (KGEs) specifically tailored for autonomous driving data. The significance of KGs in managing data stored within enterprise data lakes has been well-established in various industries. KGs have been instrumental in ensuring data findability, accessibility, interoperability, and reuse across enter-prises. The authors leveraged this established utility of KGs and investigated their application in the AD domain [12]. They used two popular benchmark datasets from Aptiv and Lyft and translated these datasets into KGs with vary-ing degrees of detail. These KGs were then transformed into KGEs using three renowned embedding algorithms: TransE, RESCAL, and HoLE. The evaluation framework adopted by the authors was comprehensive, considering four dimen-sions: quality metrics, KG informational detail, KGE algorithms, and datasets. The categorization measure stood out among the metrics used for evaluation as it captured how well entities "typed" by the same background concept clustered together. However, the coherence measure was less meaningful in this domain. The results of their study were insightful. They found that KGs with the high-est levels of informational detail were more effective in capturing both type and relational semantics than their less expressive counterparts. This finding un-derscores the importance of the depth and detail of information in KGs when generating embeddings. In conclusion, the research by Wickramarachchi et al. provides valuable insights into the potential of KGs and KGEs in the AD indus-try. Their systematic approach to evaluating KGEs offers a blueprint for future research.

Chowdhury et al. (2020) presented Autonomous Driving (AD) has seen sig-nificant advancements, leading to the creation of large knowledge graphs (KGs) that store factual data about road scenes. These KGs complement visual and decision-making systems. However, they lack commonsense knowledge, which is crucial for sophisticated decision-making in AD scenarios. Existing KGs for AD primarily store factual data, but they lack commonsense assertions that describe comparative properties between objects [13]. The paper's goal is to enrich AD KGs with commonsense knowledge. The authors have enhanced an existing KG for AD by linking it to assertions from four popular domain-agnostic common-sense knowledge repositories. The paper studies the effect of this enhancement on scene entity prediction and explainable scene clustering. S. Chowdhury, R. Wickramarachchi, M. Gad-Elra et al identified a set of road entities and ex-panded them using synonyms from WordNet. These entities were then used to collect an AD-specific slice from three existing commonsense KGs. The col-lected data was then refined to remove synonymous relations. The proposed AD-commonsense KG was integrated into an existing AD KG constructed from the PandaSet dataset.The benefits of integrating commonsense knowledge into AD KGs were evaluated through two tasks: scene entity prediction and explain-able clustering. The results showed that KG embeddings learned with integrated commonsense knowledge outperformed those without it. The integration of com-monsense knowledge into AD applications is crucial. The preliminary results of this study demonstrate the importance of integrating commonsense knowledge in AD applications. The constructed commonsense-enriched KG was found to be useful for scene entity prediction and explainable clustering.

Kontopoulos et al. (2021) introduced A major challenge lies in effectively fusing the large volumes of diverse sensory data generated by various sensors on the vehicle itself (cameras, LiDAR, RADAR, etc.) and contextual data from the environment (weather, trafic, other vehicles). Ontologies and semantic technolo-gies are well-suited to address this challenge by providing a way to semantically integrate and fuse heterogeneous data sources into a unified knowledge represen-tation model called a Knowledge Graph. They present an extensible semantic data fusion framework called CASPAR for autonomous vehicles; part of a larger platform developed in an EU project. CASPAR has a microservice architecture with different monitoring components that collect and analyze data related to the driver, vehicle, and surroundings. It ingests the analysis outputs from these components and maps the data fields to ontology concepts to populate a domain-specific Knowledge Graph. The SOSA ontology is the core semantic model for representing sensors, observations, and measured properties. They present two scenarios to demonstrate and evaluate CASPAR's capabilities: Comparing the robustness of different odometry algorithms (visual, LiDAR, cooperative local-ization) under varying conditions by analyzing their ATE and RPE metrics en-coded in the KG. Calculating driver drowsiness risk levels by fusing driver mon-itoring and road occupancy data in the KG and applying rules. The Knowledge Graph representation enabled useful insights like identifying LiDAR odometry is more robust than visual in low light conditions. Rules executed on the KG can derive risk levels from driver drowsiness (PERCLOS) and road occupancy factor.

Halilaj et al. (2023) introduced Scene understanding is a key challenge in autonomous driving and requires representing the entities and relations in a scene as a knowledge graph (KG). This can be used for entity prediction to improve scene understanding. The problem of Knowledge-based Entity Prediction (KEP) is introduced to predict potentially unrecognized entities in a scene by using the knowledge of driving scenes. The methodology develops a KG of driving scenes using datasets like Pandaset and NuScenes. It defines a Driving Scene Ontology to structure the KG. KEP is mapped to link prediction (LP) by adding an ”includesType” relation between scenes and entity types. Knowledge graph embeddings (KGE) are learned for LP and reused for KEP. KEP performance is evaluated using ranking metrics like Hits@1 and a KEP accuracy metric. An association rule mining baseline is introduced for comparison .

Zhang et al. (2022) presented that Understanding novel trafic situations requires combining domain-specific and causal commonsense knowledge. Prior work focused on perception-based methods. Gaining insights by studying text-based methods for trafic understanding using two QA datasets: BDD-QA for causal reasoning and HDT-QA for domain knowledge. Adapted three knowledge-driven approaches: NLI models trained on MNLI, commonsense models with knowledge graph self-supervision, and retrieval-based QA. Constructed BDD-QA from BDD-X annotations and HDT-QA from driving test questions. Divided datasets into partitions for analysis. Evaluated models like RoBERTa trained on synthetic data from Commonsense Knowledge Graph. Give a ”reasonableness” score for each question-candidate pair. Models learn well to answer questions about types of knowledge contained in training data but don’t generalize well. W¨ormann et al. (2023) introduced an overview of techniques for combining do-main knowledge with data-driven machine-learning models to improve robust-ness, generalization, and conformity with basic principles. This knowledge inte-gration is particularly important for safety-critical applications like autonomous vehicles. Some of the key methods discussed include incorporating physics laws into training models through auxiliary losses and constraints, embedding knowl-edge graphs into neural networks through neural-symbolic integration, and using attention mechanisms to focus models on relevant inputs. This paper also ex-plores using simulations and generative models to augment datasets. State space models that encode system dynamics are another example. Reinforcement learn-ing can be steered according to an understanding of goals and environments. Deep learning can tap into prior knowledge contained in maps for perception tasks. The accumulation of knowledge across multiple tasks and domains is ex-amined through transfer learning, continual learning, and meta-learning. Rule extraction and structured prediction are evaluated for distilling patterns from data. Natural language processing may obtain expertise from legal text. Vi-sual analytics and explanation techniques like saliency maps provide human-interpretable knowledge. Finally, the conformity of models is assessed through uncertainty estimation, causal reasoning, and rule compliance checking. Real-world examples are given for applying these knowledge integration approaches to perception, situation understanding, and planning functions for self-driving vehicles. This paper comprehensively examines how domain knowledge can be combined with machine learning, focusing on safe autonomous driving applica-tions.

El Asmar et al. (2020) introduced an AWARE ontology designed to represent the situational awareness of autonomous vehicles operating in manufacturing en-vironments. The ontology builds on existing ontologies like SUMO and SSN and extends them to incorporate concepts relevant to manufacturing autonomous vehicle perceptio. The ontology has three layers - a meta-ontology layer aligning concepts with SUMO, an ontology schema layer representing domain knowledge, and an instance layer forming a knowledge graph. The environment model rep-resents dynamic assets like humans, vehicles, and storage units. The perception capability incorporates sensors, computational models, features extracted from sensor data, and observations linking them together. The ontology models both exteroceptive sensors like cameras and LiDAR as well as proprioceptive sensors like IMU and battery sensors. Features extracted include visual bounding boxes, load status, and battery information. Observations are created to link sensors, computational models, extracted features, and perceived entities. The ontology was evaluated by checking competency questions and expert reviews. The ontol-ogy can provide a basis for autonomous vehicle awareness and behavior control in manufacturing plants. Future work includes developing reasoning and policies over the ontology and grounding robot operations using the ontology.

## 7.4 Knowledge Graph for Climate Change

Wu et al. (2022a) proposed a climate knowledge graph to integrate multiple climate data sources into a linked data platform for facilitating cross-domain analysis between climate data and other domains. The knowledge graph primar-ily uses data from NOAA climate summaries, OpenStreetMap, and Wikipedia data, and supports joint SPARQL queries across these datasets. A workflow is presented to automatically retrieve and transform NOAA climate data into RDF graphs aligned with a custom ontology, and publish it as linked open data. Ge-ographic data is integrated from OpenStreetMap and linked to Wikipedia data to provide contextual information. The resulting linked data knowledge graph is deployed in a SPARQL endpoint and URIs are dereferenceable for browsing. A web interface and tutorial assist climate researchers in exploring the knowledge graph and trying sample queries. Usability testing indicates the platform suc-ceeds in providing understandable access. Key benefits include easy multi-source integration, open data enrichment, dynamic updates, and added explain ability via semantically structured data.

Wu et al. (2021) demonstrated how climate data from the NOAA database can be modeled as knowledge graphs using RDF and the Climate Analysis on-tology and stored in a SPARQL endpoint for automated climate analyses via SPARQL queries. The ontology models climate stations and observations. Two sample use cases highlight the benefit of representing climate data as knowl-edge graphs - analyzing monthly temperature variation in Dublin and Manston from 1980-2019 using statistics on quantitative data, and analyzing weather type distribution in Sculthorpe, UK from 1951-1963 using categorical meteoro-logical records. Overall, knowledge graph representation of climate data enables interoperable, semantically integrated climate data analysis through SPARQL queries. The key ideas are leveraging knowledge graphs and semantics to en-able integrated, automated climate data analysis by modeling climate data with an ontology to support SPARQL queries for climate studies according to Islam (2022).

Fotopoulou et al. (2022) developed KnowUREnvironment, a Knowledge Au-tomation Graph concentrated on climate change and environment-related issues. Well-known examples of Knowledge Graphs that demonstrate their ability to improve public understanding, identify misinformation, and support AI systems include Google's Knowledge Graph and Amazon's product graph. Nonetheless, there is hardly a knowledge graph regarding climate change. The construction of KnowUREnvironment is based on triple extraction with Abstract Meaning Representation, syntax verification, and graph creation utilizing evidence counts They evaluate a sample of 650 triples, finding 75.85 precision and high syntactic correctness and understandability. KnowUREnvironment represents knowledge as RDF triples with subjects, predicates, and objects, and the text corpus and Knowledge Graph are publicly available to support further natural language pro-cessing and education on climate change. Overall, this work demonstrates the feasibility of automatically constructing a Knowledge Graph for climate change, and if expanded, KnowUREnvironment could become an asset for researchers, educators, policymakers, and the public seeking to understand this complex and important domain .

Chen et al. (2020) illustrated the concept of knowledge graphs KGs (Knowl-edge Graph) is considered in this work dedicated to following up on the achieve-ment of sustainable development goals SDGs. This is due to their capacity to link data warehouses, create meaningful data models, and assess complex sys-tems such as the SDGs that are intricate. KGs have been said to manage infor-mation volatility, provide semantics, and support reasoning; even so, they still face problems with their data quality, interoperability, and accessibility. This is a methodology that incorporates idea generation, development, and use of Sus-taingraph – a KG for following up SDGs with Neo4j and GraphML approaches [23]. The labeled-property graph model consists of interlinked components com-prising such entities as the SDGs, indicators, policies, hazards, innovations, and the case studies are incorporated in SustainGraph. In a continuous process that uses global and regional indicators, socio-economic aspects of data population, case studies, reports, and papers. Although not included here, the possibility of computer vision and image is acknowledged. In summary, SustainGraph illus-trates the usefulness of the KG paradigm for ongoing, adaptable, and knowledge creation for sophisticated sustainability realms.

Mishra and Mittal (2021) provided an intensive process used in building a knowledge graph for typhoon intensity improvement. Datasets of meteorological parameters for wind speed, storm central pressures, and track were meticulously prepared from 1317 typhoons over six decades. To obtain a complete under-standing of this disaster, we considered data about the economics, number of fatalities, and location of incidents taken from national and provincial yearbooks while incorporating information on infrastructure and cities from various online databases. The diverse and heterogeneous dataset was subsequently combined with an ontology-based knowledge graph and RDF triples; in this case, links were established to several types of typhoon events, affected cities, infrastructures, and disaster statistics. In this case, the study involved a prediction component, which implemented various machine learning approaches like regression, SVMs (support vector machines), and neural networks. The unique characteristic of this methodology is the use of knowledge graph data to enrich the predictive models. As a first step, supplementary input features in the form of disaster statistics and location data were introduced into the models providing them essential contextual information. Also, there are knowledge graph embeddings which are expressed as vectors using strategies such as TransE and Distrust. The embeddings were then applied in the training stage for the RNN model. This pre-training is important, as it provides the model with some relational knowledge for all entities within the knowledge graph. For example, embedding a specific city covered its spatial characteristics and the history of previous typhoons, existing infrastructure, and casualty records. Relational knowledge was vital be-cause it helped in improving the understanding of how weather parameters in-fluence different events. The pre-trained model, fine-tuned on the meteorological data, demonstrated substantial performance gains, highlighting a reduction in intensity prediction errors by an impressive 23-31 percent compared to baseline models. The recurrent neural network model, particularly when initialized with knowledge graph embeddings, emerged as the most effective in leveraging rela-tional knowledge for enhanced typhoon intensity prediction. This multi-faceted approach, integrating diverse data into a knowledge graph and utilizing graph embeddings for pre-training, presents a nuanced and robust framework for ad-vancing predictive capabilities in the realm of natural disaster forecasting.

Ge et al. (2022) explained a way to develop a knowledge graph (KG) on climate change, Ge et al. (2022) suggested using an end-to-end neural named entity relationship extraction model known as NeuralNERE for climate change knowledge graph (KG) construction, directly from the raw text of relevant news articles. This method replaced human intervention, which was expensive during supervised and semi supervised KG construction. A new climate dataset was introduced called SciDCC dataset which was scraped from the Science Daily website where NeuralNERE can be utilized to gather historical data to produce climate change knowledge graphs. Learning the embedding representation of the intended relationship phrase, which characterizes the relationship between any two named entities in the previously presented SciDCC dataset, is the main goal of the NeuralNERE model. NeuralNERE model Using the raw text extracted from the body and summary sections of every article in the SciDCC dataset, we first generate an input corpus. The preprocessing of this input corpus includes tokenization, lower-casing, stemming, and lemmatization. This corpus is then used for the fine-tuning of a character-based language model like FastText or GloVe to learn the representations of the word embeddings for each word in the corpus for obtaining every named entity phrase and every potential connection phrase. Part-of speech tagging is used to extract the named entity phrases and potential connection phrases (POS Tagging). Every word token has a POS tag associated with it. These tags are used to extract all named entity phrases using noun-phrase chunking, and all potential relationship phrases using verb-phrase chunking, to produce two lists: an entity phrase list and a relationship phrase list. The extracted named entity phrases from the entity phrase list and the re-lationship phrases from the relationship phrase list are then translated into their appropriate embedding representations using the fine-tuned language model ac-cording to Mishra and Mittal (2021). To prevent issues while creating embedding representations for multi-word entity, and relationship phrases, we suggest using a character based language model. The relationship between the subject entity phrase (s) and the object entity phrase (o) is encoded into an encoded represen-tation zij in R by NeuralNERE using an encoder decoder network. This encoded representation has the same embedding representation size as the relationship phrases. The main results found were Gathering and classifying data on climate change from a vast library of news stories, examining the connections between the many causes of climate change, and Compiling analysis and reasoning on the key incidents to enable better-informed policymaking regarding climate change.

Wu et al. (2022b) explained that In real-time, remote sensing technology can quickly and eficiently gather vast amounts of highly dynamic data over ex-tended observation ranges. It combines artificial intelligence and remote sensing technologies to extract information about cultivated land, buildings, and other things from the ground, including fine details. In the areas of early warning, emergency decision making, and catastrophe monitoring, it is crucial. When ex-tracting remote sensing information for multi-scale, multi-temporal, multi-type, multi-precision remote sensing extraction information, other geographic infor-mation, and other spatio-temporal data in disaster scenes, the crucial issue is the fusion of various data types, different data structures, and diverse types of expert knowledge. The hierarchical semantic model of disaster prediction and the mapping connection between disaster prediction scenarios and DPKG archi-tecture are the primary topics covered in this section’s theoretical explanation of how disaster prediction scenarios are mapped to knowledge graphs. The dis-aster prediction scenario must be represented as a hierarchical semantic model to model catastrophe prediction using a knowledge graph. It uses the structure of the semantic model to transfer the spatiotemporal features of the data in the catastrophe prediction scenario into the DPKG. One may turn the catastrophe analysis model into knowledge. To anticipate disasters, the DPKG must addi-tionally incorporate expert knowledge and the modified disaster analysis model. DPKG architecture consists of an ABox fact collection and a TBox rule set. The conceptual layer (a shared semantic ontology for catastrophe prediction) and inference methods are contained in TBox. The instance layer of ABox in-cludes spatiotemporal and attribute data on things associated with catastrophes. Makes use of OWL to represent knowledge. Space ontology, temporal ontology, and common semantic ontology are all part of the conceptual layer. First-order logical rules, production rules using SWRL (Semantic Web Rule Language), and spatiotemporal semantic rules are examples of inference rules. Knowledge is ex-tracted by the instance layer from semi-structured topography and geology data, organized meteorological data, and unstructured remote sensing data. includes models for catastrophe prediction as well. DPKG enables spatiotemporal seman-tic data searching using SPARQL. First-order logic, production rule, and spa-tiotemporal semantic reasoning are examples of reasoning techniques. Dynamic data-driven prediction is made possible by chaining rules using spatiotemporal reasoning. The DPKG for geological landslide and forest fire prediction case studies is demonstrated in the study. In conclusion, it’s clear that the suggested DPKG technique could integrate data from several sources, portray models as rules of reasoning, facilitate temporal and spatial reasoning, and facilitate dy-namic catastrophe prediction. The outcomes confirmed the advantages in terms of performance, automation, and forecast accuracy.

Wu et al. (2023) focused on addressing the data isolation problem in climate analysis by leveraging semantically uplifted knowledge graphs. The authors high-light that climate data resources available on the internet are often fragmented and stored in different formats, making it challenging to integrate and analyze the data effectively. To overcome this challenge, the authors propose the use of knowledge graphs, which are constructed based on a semantic model and facili-tate data interoperability. They specifically target the Python machine learning environment and emphasize the integration of remote climate knowledge graph data with local tabular data, particularly in the form of CSV files. The research utilizes a variety of climate data sources, including station-based sensor data. These sensor data may come from various sources and be stored in formats such as CSV, JSON, and XML. The authors aim to integrate these disparate data sources into a unified knowledge graph, enabling seamless access and analysis. The authors developed an open climatic knowledge graph (KG) that follows linked data principles. This KG provides online access to multisource climate data, covering various aspects such as weather, atmospheric conditions, and air quality. By adhering to linked data principles, the KG ensures better data acces-sibility and facilitates connections with other linked data sources. In their case study, the authors focus on rainfall detection using the NOAA (National Oceanic and Atmospheric Administration) tabular data. They demonstrate how incor-porating the additional climate knowledge graph data can enhance the machine learning model’s performance in predicting and classifying rainfall. The results show an improvement in accuracy, with up to a 10 percent increase, compared to using only the original tabular data.

Web (2020) focuses on the creation of a global Climate Action Knowledge Graph to accelerate progress on climate change. The paper highlights the im-portance of advanced data and information management practices in support-ing climate action and sustainable development. It introduces the concept of knowledge graphs, which are tools that connect communities and agendas, fa-cilitate learning, and enable collaboration. The paper proposes the development of a comprehensive Climate Action Knowledge Graph that interconnects data, information, people, organizations, and businesses globally to implement effec-tive climate policies and measures. The use of artificial intelligence (AI) is em-phasized, as it can leverage the knowledge graph and contribute to innovation and progress. The paper discusses the potential benefits of the Climate Action Knowledge Graph, such as rapid knowledge discovery, eficient exploration of connections, improved learning approaches, and enhanced collaboration. It also highlights the role of AI technologies in supporting climate action, sustainable development, and resilience. The paper concludes by emphasizing the need for innovative information technologies and knowledge management systems to ad-dress climate change challenges effectively.

Wu et al. (2023) assert that while the impact of climate change on various industries has been well-documented, its effect on the tourism economy is not yet fully understood. They explore the use of knowledge graph techniques ap-plied to naturalistic data, particularly weather data, to enhance understanding of how a country's tourism economy and climate interact. The paper proposes a "Knowledge Graph-based Tourism Analytic System". This system employs knowledge graph techniques such as RDF (Resource Description Framework) and SPARQL (SPARQL Protocol and RDF Query Language) to organize cli-mate and tourism data eficiently. They aim to provide users with improved insights into location, lodging, and local attractions. The model used in this pa-per is a knowledge graph, a powerful tool for modeling real-world entities and relationships and capturing semantic meaning. The authors note that knowledge graphs are particularly useful for obtaining intelligence through artificial intel-ligence and machine learning. The data used in this paper comes from multiple sources. The climate data, which includes rainfall, temperature, and wind data, is primarily collected from North America. This is integrated with tourism data, such as flight data, to create a multidomain ontology. Some of the data sources include the National Oceanic and Atmospheric Administration (NOAA), Avia-tionStack, and Simple Maps. The authors have created a multidomain ontology to integrate climate data and tourism data from various sources. The ontology allows for the organization of diverse sources of data in a way that is reusable and adaptable to schema modifications. This enables the creation of a climate-tourism knowledge graph platform that improves tourism analytics by taking cli-mate into consideration. The semantic schema of the proposed climate-tourism knowledge graph has been published on the Web for potential reuse and schema sharing across heterogeneous data sources. Overall, the paper combines an anal-ysis of challenges, an introduction to knowledge graphs, practical examples, and a proposal for the development of a global Climate Action Knowledge Graph to address knowledge gaps and accelerate climate action. The paper suggests the development of a global Climate Action Knowledge Graph, which would inter-link disconnected data, information, people, organizations, and businesses. This implies that the data involved would encompass a wide range of climate-related information, such as scientific research, policy documents, reports, datasets, case studies, and potentially other relevant sources. The primary goal of the proposed Climate Action Knowledge Graph is to leverage this interconnected data to fa-cilitate knowledge discovery, enhance collaboration, and support evidence-based decision-making in the context of climate change and sustainable development.

Yu et al. (2022) published in Geoscience Frontiers in 2023, presents a study on the development of a climate paleogeography knowledge graph, specifically focusing on deep time paleoclimate classifications. The authors, Chenmin Yu, Laiming Zhang, Mingcai Hou, Jianghai Yang, Hanting Zhong, and Chengshan Wang, aim to enhance the understanding and comparison of global and regional climate changes by unifying scientific concepts from different paleoclimate clas-sifications. The authors note that due to the qualitative constraints and incon-sistent descriptions of climate types, the application of climate classification in deep time (i.e., climate paleogeography) is challenging [29]. To overcome this, they established a climate paleogeography knowledge graph under the frame-work of the Deep-Time Digital Earth program (DDE). This hierarchical knowl-edge graph consists of five paleoclimate classifications based on various strategies. They fully evaluate the strengths and weaknesses of these classifications in terms of simplicity, applicability, quantifiability, and comparability. The data for their research is primarily qualitative, derived from published paleoclimate classifica-tions and climatic proxies. They also reconstruct the global climate distributions in the Late Cretaceous according to these classifications, comparing the results and evaluating the relationships among these climate types in different classifica-tions. A knowledge graph model to represent and analyze the data. This model represents data as a graph with nodes and edges, where nodes represent entities and edges represent relations between these entities. This approach is useful for standardizing database format, promoting data interoperability, and addressing the issue of heterogeneity in paleoclimatology research. The data used in this study is derived from various sources, including existing paleoclimate classifica-tions, climatic proxies for deep time climate, and the Deep-Time Digital Earth program (DDE) [29]. The authors also used the global paleoclimatic distribu-tions during the Late Cretaceous for their analysis. The use of SWEET ontology (Semantic Web for Earth and Environmental Terminology), which is a major Earth Science ontology covering over 200 disciplines and 6000 concepts in earth and environmental science. It also references the Geolink project, which aims to handle heterogeneous problems caused by differences in data formats and meth-ods of access. The authors suggest that similar ontological approaches would be beneficial in establishing a climate paleogeography knowledge graph.

## 8 Conclusion

Knowledge graphs are really improving how data is used these days. By con-necting different types of information together and learning from examples, it helps integrate, understand, and apply data in many fields much better. Systems that use knowledge graphs can adapt more easily to new situations because they combine rules about specific topics with patterns found in data. This allows knowledge graphs to build very thorough models of things and how they relate to each other.Finding ways for knowledge graphs to eficiently manage and link together even larger sets of data with more complex webs of connections is a big focus. In addition, ensuring these models are understandable will help en-sure they can be trusted for serious applications. By continuing advancements, knowledge graphs will play an increasingly important job in helping many in-dustries gain benefits from massive amounts of information. As knowledge graph technologies continue to evolve, they will further enhance our ability to leverage data, ultimately driving innovation and eficiency across various domains.